\documentclass{article} 
\usepackage{asstyle}

\usepackage{microtype}
\usepackage{hyperref}
\usepackage{url}
\usepackage{booktabs}
\usepackage{todonotes}
\usepackage{multirow}
\usepackage[whole]{bxcjkjatype}

\usepackage{tcolorbox}
\usepackage{fancyvrb}
\usepackage{CJKutf8}
\usepackage[frozencache=true,cachedir=minted-cache]{minted}
\tcbuselibrary{breakable}
\tcbuselibrary{skins}

\newenvironment{promptbox}[2]{%
  \begin{tcolorbox}[
    breakable,
    enhanced,
    colback=gray!5,
    colframe=gray!50!black,
    arc=0mm,
    title={#1},
    fonttitle=\bfseries,
    boxrule=0.5pt
  ]
  \begin{CJK}{UTF8}{min}
  \begin{flushleft}
  \small
  \VerbatimInput[fontsize=\small,breaklines=true]{#2}
  \end{flushleft}
  \end{CJK}
  \end{tcolorbox}
}{}

\definecolor{darkblue}{rgb}{0, 0, 0.5}
\hypersetup{colorlinks=true, citecolor=darkblue, linkcolor=darkblue, urlcolor=darkblue}

\newcommand{\vendordf}{Vendor~1}
\newcommand{\vendorfl}{Vendor~2}
\newcommand{\TransEvalnia}{TransEvalnia}

\title{\TransEvalnia: Reasoning-based Evaluation and Ranking of Translations}

\author{Richard Sproat, Tianyu Zhao \& Llion Jones\thanks{ Sakana.ai \\
    Toranomon Hills Business Tower\\
    1-17-1 Toranomon \\
    Minato City, Tokyo \\
    105-6415 \\
    Japan \\
    \texttt{\{rws,tianyu,llion\}@sakana.ai} }
  }

\begin{document}

\maketitle

\begin{abstract}
We present \TransEvalnia, a prompting-based translation evaluation and ranking
system that uses reasoning in performing its evaluations and ranking. This
system presents fine-grained evaluations based on a subset of the
Multidimensional Quality Metrics (\url{https://themqm.org/}), returns an
assessment of which translation it deems the best, and provides numerical scores
for the various dimensions and for the overall translation. We show
that \TransEvalnia\ performs as well as or better than the state-of-the-art
MT-Ranker \citep{Moosa:EtAl:24} on our own English$\sim$Japanese data as
well as several language pairs from various WMT shared tasks. Using Anthropic's
Claude-3.5-Sonnet and Qwen-2.5-72B-Instruct as the evaluation LLMs, we show that
the evaluations returned are deemed highly acceptable to human raters, and that
the scores assigned to the translations by Sonnet, as well as other LLMs,
correlate well with scores assigned by the human raters. We also note the
sensitivity of our system---as well as MT-Ranker---to the order in which the
translations are presented, and we propose methods to address this
\emph{position bias}. All data, including the system's evaluation and
reasoning, human assessments, as well as code is released.
\end{abstract}

\section{Introduction}
\label{sec:introduction}

Translation systems using large language models (LLMs) have become so good that
such systems can even beat human translations on some tasks
\citep{Freitag:EtAl:24}, and reasoning abilities of LLMs are improving
performance in many areas including document-level translation and literary text
\citep{Liu:EtAl:2025}. As the quality bar is raised, it becomes more difficult
to improve over previous systems, but at the same time it also becomes more
important to have reliable automated evaluation methods.

Automated tools for evaluating MT systems have of course been around for more
than two decades. The classic BLEU metric \citep{Papineni:EtAl:02}, while widely
criticized for years \citep{Callison:EtAl:06,Mathur-EtAl-20}, is still widely used as a
cheap and often effective way to tease apart different translation systems. Many
more sophisticated metrics exist including MetricX \citep{Juraska:EtAl:23},
including systems that provide a ranking for a pair of translations
\citep{Moosa:EtAl:24}.

The main drawback of these rating systems is that they provide a numerical
score, but typically do not present any reasons for the scoring. As the quality
of machine-generated translations improves, and as such translations become
increasingly hard to separate from human translations, the value of simply
having a raw number or ranking decreases. Rather it becomes increasingly
necessary to have a system that can present detailed reasoning behind a rating
or ranking. Furthermore, it is desirable for the system to produce reasoning
along various dimensions such as how \textbf{accurate} the translation is, how
correct the \textbf{terminology} is and how \textbf{appropriate it is for a
  target audience}, dimensions that are supported by detailed translation
evaluation rubrics such as Multidimensional Quality Metrics
(MQM\footnote{\url{https://themqm.org/}}). If the evaluation system presents its reasons for
the evaluations, then the user can evaluate the evaluations, and understand why
the system came to the conclusion it did. The user can also more easily identify
weak points, and cases where the reasoning is misguided and perhaps come to a
different conclusion as to which translation to prefer.

In this paper we propose \TransEvalnia, a prompting-based translation evaluation
and ranking system that presents fine grained evaluations for a subset of the
dimensions of the MQM, can rank multiple translations on the basis of those
evaluations, and can assign scores on a 5-point Likert scale to the target
translations along the various dimensions, as well as an overall score for the
translation.

Our main focus is on English$\sim$Japanese translation, where we evaluate our
system on a range of genres ranging from news texts, where automated translation
systems already perform very well, through more difficult genres such as
proverbs, and haiku. In addition we evaluate our evaluation system on various
WMT datasets---English-German, Chinese-English, English-Japanese,
Japanese-English, English-Russian, English-Spanish where expert human MQM-style
scores are available, and compare our system with the state-of-the-art MT-Ranker
system reported in \cite{Moosa:EtAl:24}. We show that for all datasets, except
WMT-2023~en-es, \TransEvalnia\ is on a par with or outperforms MT-Ranker.

All code and data---for details, see Section~\ref{sec:data} and
Appendix~\ref{subapp:data}---is
open-sourced.\footnote{Code at \url{https://github.com/SakanaAI/TransEvalnia} and data at \url{https://huggingface.co/datasets/SakanaAI/TransEvalnia}}.

\section{Previous evaluation methods}
\label{sec:background}

The classic BLEU metric~\citep{Papineni:EtAl:02}, despite being widely used to assess machine translation systems, has shown diminishing utility as contemporary MT systems produce translations with quality comparable to, or surpassing human translations~\citep{Callison:EtAl:06, Mathur-EtAl-20}.

BLEURT~\citep{BLEURT} introduced a new paradigm that fine-tunes pretrained models for encoding and evaluating text generation systems. In machine translation specifically, the COMET series~\citep{CometKiwi, Guerreiro:EtAl:23:comet} and the MetricX series~\citep{Juraska:EtAl:23, MetricX24} represent state-of-the-art approaches that have consistently achieved superior results in the WMT shared tasks for metrics and quality estimation. Both model families fine-tune pretrained transformers—XLM~\citep{XLM} and mT5~\citep{mT5}, respectively—employing carefully designed data collection and training strategies.

Recent frontier LLMs such as GPT~\citep{gpt3} demonstrate exceptional performance in translation evaluation through zero-shot or few-shot prompting. GEMBA~\citep{GEMBA} prompts GPT variants to directly output numerical rating scores.~\citet{Fernandes:EtAl:23} prompt PaLM2 to generate structured evaluations following the MQM framework. Similarly, INSTRUCTSCORE~\citep{Xu:EtAl:23} leverages GPT-4 to produce MQM-style structured assessments. MT-Ranker~\citep{Moosa:EtAl:24} achieves state-of-the-art performance across multiple metric datasets by implementing a pairwise evaluation methodology. Notably, pairwise evaluation not only enhances model performance but also significantly improves agreement in human annotation compared to pointwise evaluation approaches~\citep{Song:EtAl:25}.

LLMs possess the inherent capability to explain their decision-making processes~\citep{Anker:EtAl:24}, and soliciting rationales from these models has been shown to improve decision quality. \citet{Yan:EtAl:24} demonstrated that structured comparative reasoning outperforms both direct comparison and Chain-of-Thought (CoT) baselines in general text preference tasks. Reason-based evaluation approaches can also enhance subsequent refinement of decisions~\citep{LLMRefine}. Despite these advances, reason-based evaluation remains largely unexplored specifically for assessing translation quality.

\section{LLMs}
\label{sec:llms}

Table~\ref{tab:llms} lists the LLMs we have used for various parts of this
project, giving the full specification of the LLM, and the abbreviated name that
we will use in the discussion.

\begin{table}[t]
\begin{center}
\begin{tabular}{lr}
\toprule
Abbreviation & Full specification\\
\midrule
Mistral & mistralai/Mistral-7B-Instruct-v0.3\\
Qwen & Qwen/Qwen2.5-72B-Instruct-GPTQ-Int4\\
Llama & unsloth/Llama-3.3-70B-Instruct-bnb-4bit\\
GPT-4o & GPT-4o-mini\\
GPT-4o-24 & GPT-4o-2024-08-06\\
GPT-4 & GPT-4-1106-preview\\
Sonnet & us.anthropic.claude-3-5-sonnet-20240620-v1:0\\
\bottomrule
\end{tabular}
\end{center}
\caption{\label{tab:llms}LLMs used in this work}
\end{table}

\section{Data}
\label{sec:data}

We summarize our data in this section. For full details see
Appendix~\ref{subapp:data}.

\subsection{Data without human rankings}
\label{subsec:data_noranking}

Two datasets have no human rankings of the translations. Our first dataset
consists of 862 English proverbs from Eigokotowaza\footnote{\url{https://eigokotowaza.net}} with
reference Japanese translations to which we added translations from Mistral,
Qwen, Sonnet and GPT-4o-24.  This was combined with 1,997 English news sentences
from NTREX, again with reference Japanese translations, and LLM-generated
translations.  We selected 50 examples of an English source and its Japanese
translations, from each of the above sets and combined them into a dataset which
we will henceforth term \textbf{Generic}. The motivation is to provide a set of
translations that ranges in difficulty from news style texts---relatively easy to
translate; to proverbs---harder to translate, see e.g. \citep{Wang:EtAl:25}.

Our second dataset consisted of 1,065 haiku by Matsuo Bash\=o (1644--1694), some
with English translations. In addition to the few cases with reference
translations, we also generated LLM translations using Mistral, Qwen, Sonnet and
GPT-4.  From these we generated a random subset of 100 with the original
Japanese source and all translations (henceforth \textbf{Haiku~100}), in
addition to using the full set (\textbf{Haiku~Full}).

\subsection{Data with human rankings}
\label{subsec:data_ranking}

Our \textbf{Hard~en-ja} dataset consists of 10 English source sentences with
multiple automated translations from Google Translate, GPT 3.5, GPT 4,
and GPT with a reasoning phase. Two human experts then rated the translations on
a 10-point scale.  In addition we used datasets from WMT-2021
English-Japanese/Japanese-English, WMT-2022 English-Russian, WMT-2023
Chinese-English and English-German and WMT-2024 English-Spanish. These provide
multiple translations with human scores for the translation allowing for
ranking. From each of the WMT datasets we randomly selected 500 examples.

\section{Methods}
\label{sec:methods}

\subsection{Evaluation criteria}
\label{subsec:evaluation_criteria}

Given a text in the source language, and two or more translations into a target
language, our task is to provide a detailed evaluation of each translation, and
determine which translation should be considered the best. Since a central task
is evaluation, we start with a description what kind of evaluation should be
returned by the system.

We adopt a small set of key evaluation dimensions that are based on a subset of the
Multidimensional Quality Metrics. In particular we
consider the following dimensions:
\begin{enumerate}
\item
ACCURACY: Does the translation convey the sense of the original accurately?
\item
TERMINOLOGY: Do the terms used conform to normative terminology standards and
are the terms in the target text the correct equivalents of the corresponding
term in the source text?
\item
LINGUISTIC CONVENTIONS: Is the translation fluid and grammatical?
\item
AUDIENCE APPROPRIATENESS: Are the chosen words and expressions familiar to a
\textit{target-language}-speaking audience?
\item
HALLUCINATIONS: This portion of the translation does not appear to correspond to
anything in the original and cannot be justified by any need to adapt the text
to the target audience. It seems like a hallucination.
\item
MISSING CONTENT: Is there any important information in the original that is
missing from the translation?
\end{enumerate}
The hallucination dimension is particularly relevant for LLM-generated
translations: human translators are particularly unlikely to produce such
output.

Since haiku are poetry, we used a slightly different set of evaluation criteria
than for other texts. In particular, while the dimensions ACCURACY, MISSING CONTENT,
HALLUCINATIONS and AUDIENCE APPROPRIATENESS were the same as for other texts,
LINGUISTIC CONVENTIONS was replaced by EMOTIONAL CONTENT, something that is
highly salient when it comes to translations of artistic language:
\begin{quote}
The EMOTIONAL CONTENT of the original: does it convey the emotive content of
   the Japanese original?
\end{quote}

\subsection{Reason-based evaluation and ranking}
\label{subsec:reason_eval}

With the criteria above in mind, we turn to the evaluation system itself.
Again, given a set of candidate translations of a source-language text into a
target language, the basic tasks we want to perform can be enumerated as
follows:
\begin{enumerate}
\item\label{it:eval} Break down each translation into one or more spans, and
  then for each span, evaluate the translation along the dimensions of
  \textbf{accuracy}, \textbf{terminology}, \textbf{linguistic~conventions},
  \textbf{audience~appropriateness}, \textbf{hallucinations} and
  \textbf{missing~content}. After evaluating the individual spans, provide an
  overall summary evaluation for the entire translation.
\item\label{it:rank} Based on the set of evaluations for each translation, provide an
  assessment of which translation is best, and give reasons for this assessment.
\item\label{it:score} Score each translation (segment) along the dimensions of
  \textbf{accuracy}, \textbf{terminology}, \textbf{linguistic~conventions},
  \textbf{audience~appropriateness}, \textbf{hallucinations} and
  \textbf{missing~content}, using a 1--5 Likert scale where `1' is worst and `5'
  is best. Provide an overall score for the entire translation, again using a
  Likert scale.
\end{enumerate}
In Step~\ref{it:eval}, the motivation for breaking down a translation into spans
is that, particularly in long texts, it may be that some portions of the
translation are overall better than others---cf.
\cite{Fernandes:EtAl:23}. This then allows for more fine-grained feedback. Thus
the first task that the LLM is required to perform is to decide a sensible set
of cut points in the translation, before it proceeds to evaluate each cut point
along the various dimensions.

Steps~\ref{it:rank} and \ref{it:score} are independent of each other: one can
either ask the system to rank a given translation as best, or provide more
detailed scoring.  For scoring, if one wants to then rank a pair of
translations, we can simply use the arithmetic mean of the scores. This is the
method used in the evaluations reported below. Since scoring applies to a given
translation and its evaluations independently of any other translation, the
scoring phase can be conducted as a separate step. Scoring-based ranking is thus
not sensitive to position bias (see below), but it also turns out to be overall
weaker than other methods.

For ranking, one can in principle combine Steps~\ref{it:eval} and \ref{it:rank}
into one step where the model is prompted first to evaluate the translations,
then consider all the translations and the evaluations taken together to
determine the best translation.  However one problem with this is position bias:
the LLM is sensitive to the order in which the translations are presented.
Position bias is a well-known problem with LLMs that affects, for example,
constraint application \citep{Zeng:EtEl:25}, as well as ranking tasks.  For
example, \cite{Ye:EtAl:24} provide a systematic study of position bias in
LLM-as-Judge tasks. They note, for example (page~8) that ``[position] bias
becomes more pronounced as the number of answers increases, particularly when
evaluating three or four options, resulting in a decreased robustness rate'',
though as we will see below, position biases are a problem even when there are
only two alternatives. Also, though \cite{Ye:EtAl:24}'s study found that Claude
Sonnet was the most robust of the LLMs they compared, in our experience
as we will see below, Sonnet is still sensitive to position, in accord with
\cite{Shi:EtAl:24}, who found that Claude-3 models ``displayed a tendency
to prefer more recent responses'' (p.~2).
\begin{figure}[t]
\begin{center}
\includegraphics[width=0.65\textwidth]{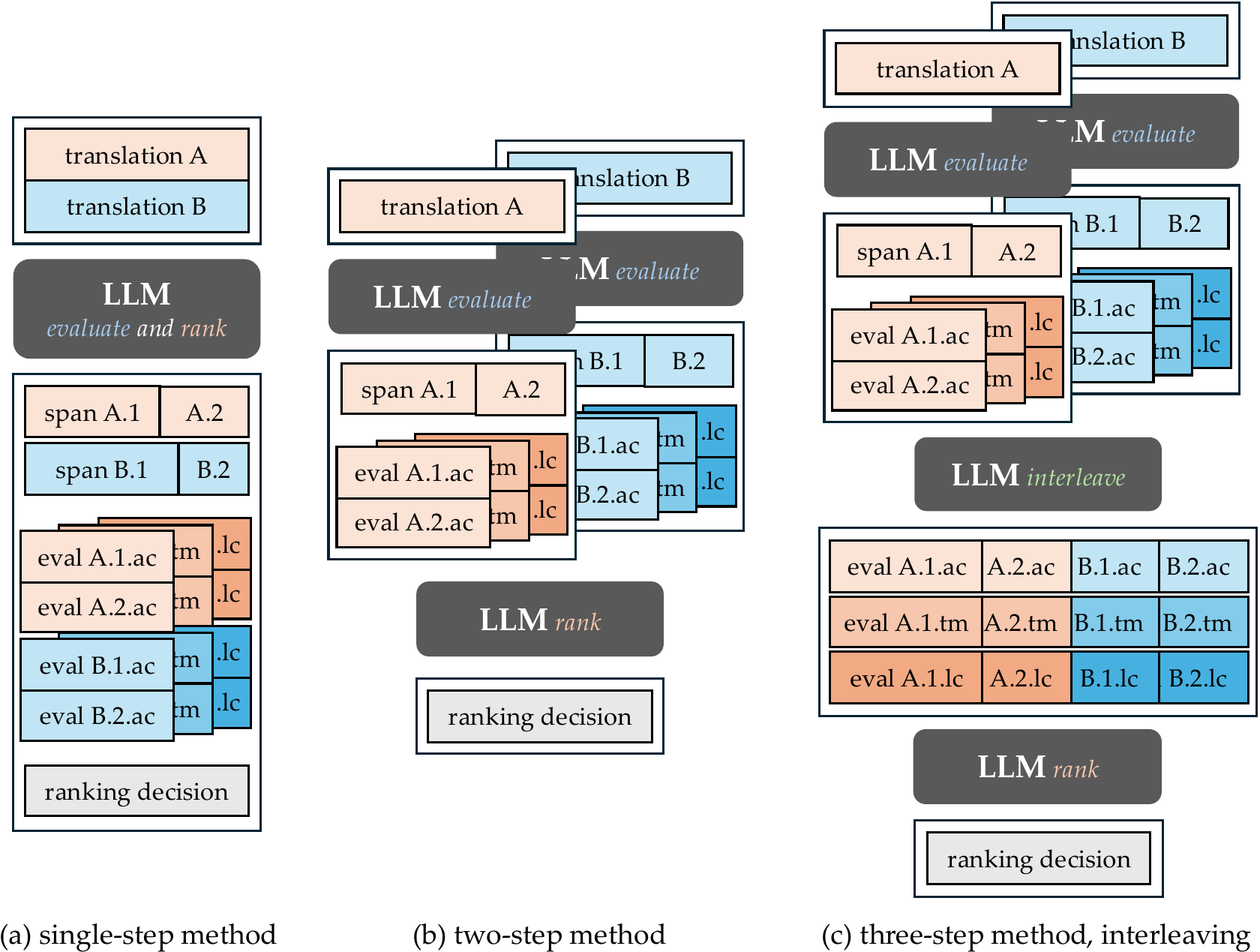}
\end{center}
\caption{\label{fig:3models}Three approaches to evaluation. (a) Single step
  where the LLM evaluates all the translations, then ranks. (b) Two step, where
  evaluations are done separately then combined and ranked. (c) Three step,
  where the evaluations are done separately, then interleaved and finally
  evaluated. In all cases expressions of the form 'eval.t.n.d' where $t \in \{A,
  B, \ldots\}$, $n \in \{1, 2, \ldots\}$, $d \in \{\textbf{ac}\textnormal{curacy}, 
  \textbf{t}\textnormal{er}\textbf{m}\textnormal{inology}, 
  \textbf{l}\textnormal{inguistic } \textbf{c}\textnormal{onventions}, \ldots\}$, 
  denote the evaluation of dimension $d$ of the $n$-th span of the $t$-th translation. 
  The actual dimensions---e.g. accuracy, terminology, etc.--- are represented as two 
  letters here for compactness.}
\end{figure}

The position bias problem of the \textbf{one-step} approach can be mitigated
somewhat by splitting the task into two separate stages, first evaluation, then
ranking. In the one-step approach, the prompt includes the source text and the
translations, and it is asked to produce an evaluation then a ranking. In the
\textbf{two-step} approach, the evaluations of each translation are done
separately, with the system only seeing one translation at a time. Then, in the
second, ranking, phase the translations along with their evaluations are
presented to the system.  Thus the LLM sees more content, which could in
principle dissuade it from being sensitive to whichever translation comes first
or last. In fact, as we will see below, this does indeed help reduce the
sensitivity to position.

To further mitigate against position bias we follow an approach proposed by
\cite{Li:EtAl:24}. In their task, they are interested in ranking two sets of
LLM-generated instructions for quality. For example two LLMs might generate
instructions for alleviating stress: which set of instructions is more useful?
\cite{Li:EtAl:24}'s solution involves interleaving the instructions, so that the
first step in the instructions from LLM$_1$ is presented followed by the
comparable first step from LLM$_2$; the second step in the instructions from
LLM$_1$ is presented followed by the comparable second step from LLM$_2$; and so
on. In the case of our translation evaluation task, the situation is a bit more
complicated since we may need to interleave more than two sets of evaluations,
depending on how many translations are to be compared. First we interleave the
$n$ translations and their decompositions into spans. Next we interleave each of
the $n$ evaluations for \textbf{accuracy}; next \textbf{terminology}; and so on
for each of the dimensions up to the final overall evaluation.  These
interleaved evaluations are then passed to the third and final phase of
ranking. In general this \textbf{three-step, interleaving} approach is the most
successful at mitigating position bias.

These three different approaches are presented in Figure~\ref{fig:3models}. The
prompt for the single-step method is presented in
Appendix~\ref{subapp:single_step}; the two prompts for the two-step method in
Appendix~\ref{subapp:two_step}; and the three prompts for the three-step
interleaving method are presented in Appendix~\ref{subapp:three_step}.
In Appendix~\ref{subapp:basho_three_step} can be found an illustration of the
three-step interleaving as applied to a Bash\=o haiku translation.

It should be emphasized that position bias is by no means particular to our
LLM-based translation evaluation system. As we will see, the MT-Ranker system
\citep{Moosa:EtAl:24} is also strongly affected by the order in which the two
translations are presented, as we will see in
Section~\ref{sec:experiments_results}.

\subsection{No-reasoning scoring}
\label{subsec:no_reasoning_scoring}

Can reasoning be dispensed with entirely, instead just asking the LLM to rank a
set of translations with no further analysis? We also tried this approach, which
we will dub the \textbf{no-reasoning} approach---cf. \cite{GEMBA}. The prompt
used is shown in Appendix~\ref{subapp:no_reasoning}.  As we will see, this
approach yields good results in terms of agreement with human-evaluated
ranking. Furthermore, the LLM may also, in addition to returning the requested
ranking, profer its reasons for the ranking, though these are much more terse
than when an explicit evaluation is requested.  However, this approach suffers
from the problem that position bias is even more severe than with the one-step
approach. Furthermore, unlike our reasoning-based approach where the process can
be broken down into phases, and the results of evaluation interleaved, there is
no good way to mitigate against position bias in the case of the no-reasoning
approach: The best one could do would be to submit the evaluations in all
possible orders, and do a majority vote for the cases where one gets different
results.

\subsection{Human rating of evaluations}
\label{subsec:human_rating}

We prepared data for human meta-evaluation of our evaluation system from our
Generic English-Japanese dataset described previously in
Section~\ref{sec:data}. For each English source sentence, we evaluated each of
the five translations using our system---Step~\ref{it:eval} above---and then
scored each translation---Step~\ref{it:score}.  Since there were 100 English
source sentences, in principle this would produce 500 distinct sets of
evaluations and scores, but a few translations were
identical, so the actual number was 497. For the LLMs we used Sonnet and Qwen,
resulting in two sets of 497 English-Japanese translations with evaluations and
scores.

For the Sonnet-produced data we selected 200 randomly selected examples and sent
them to two external linguistic annotation vendors. \vendordf~is a US-based
international translation company, whereas \vendorfl~is a Japan-based
AI-infrastructure and data service. \vendordf~assigned five raters to this
task, and \vendorfl~assigned two raters, but in both cases each example was
rated by a single rater.  All raters used by the companies were professional
English$\sim$Japanese translators. Raters were not informed which translation
system (human, or any of the four LLMs) was used to produce the translation.

Raters were instructed to look at the source English text, the Japanese
translation. They then considered the individual evaluations for each of the
rating dimensions for each span, as well as the overall evaluation, and were
asked whether or not they agreed with each evaluation. If they disagreed, they
were asked to state the reason for their disagreement. Similarly, for the
evaluation for the whole translation they were asked to state whether they
agreed with the evaluation or not, and if not, the reason for their
disagreement. Finally they were asked to score the \emph{translation} (not the
evaluation) on each of the dimensions, on a five-point Likert scale. They were
also asked to score the overall translation on a five-point scale.  The complete
set of instructions sent to the raters can be found in
Appendix~\ref{subapp:rater_instructions}.

A different random selection of 200 Qwen-produced evaluations and scores were
later sent to \vendorfl, with the same instructions for labeling.

In addition, a random selection of 200 Qwen-produced evaluations and scores of
haiku were also sent to \vendorfl, with rater instructions slightly adapted for
this case. For this task the company employed a translator who was natively
bilingual and was able to judge the more subtle aspects of poetry
translation.

\section{Experiments and Results}
\label{sec:experiments_results}

\subsection{Evaluation of ranking methods on data with human-scoring}
\label{subsec:eval_ranking}

For the sets that have human scores from which we can derive a ranking between a
pair of translations (Section~\ref{subsec:data_ranking}), in addition to our own
models, we run for comparison the state-of-the-art MT-Ranker translation
evaluation system \citep{Moosa:EtAl:24}. We use their XXL model and follow the
implementation at
HuggingFace\footnote{\url{https://huggingface.co/spaces/ibraheemmoosa/mt-ranker/blob/main/app.py}}. In
addition to MT-Ranker, we also compare against COMET-22, COMET-23-XXL,
XCOMET-XXL \citep{CometKiwi, Guerreiro:EtAl:23:comet} and MetricX-XXL
\citep{Juraska:EtAl:23, MetricX24}. Note though that XCOMET-XXL and MetricX-XXL
have been fine-tuned on WMT data, so that comparison with these systems may not
be entirely fair.

For all sets we ran the various \TransEvalnia\ models using Qwen as the LLM,
with Sonnet also being used in some cases.  We summarize the results showing the
accuracies all systems for each dataset in Figure~\ref{fig:accuracies}.
Tables~\ref{tab:hard_en_ja}--\ref{tab:wmt_en_es} in
Appendix~\ref{subapp:evaluation_of_ranking_methods} give a detailed breakdown of
the results for all datasets and all systems.

For WMT-2024~en-es, the best performing system is MT-Ranker. This presumably
reflects the abundance of English-Spanish translation data:
\cite{Kocmi:EtAl:2024} note that the best English-Spanish translation systems
are now ``close to flawless'', suggesting that this language pair is
particularly well-covered in training data for automated translation.  For all
other datasets one or more of our \TransEvalnia\ configurations are on a par
with or outperform
MT-Ranker. In the case of WMT-2023~en-de, the Qwen~no-reasoning system actually
performs the best and in general the no-reasoning systems perform quite well in
comparison to their counterparts that have explicit reasoning. However as we
will discuss in the next section, the no-reasoning systems are quite sensitive
to position bias.

\begin{figure}[t]
\begin{center}
\includegraphics[width=1.0\textwidth]{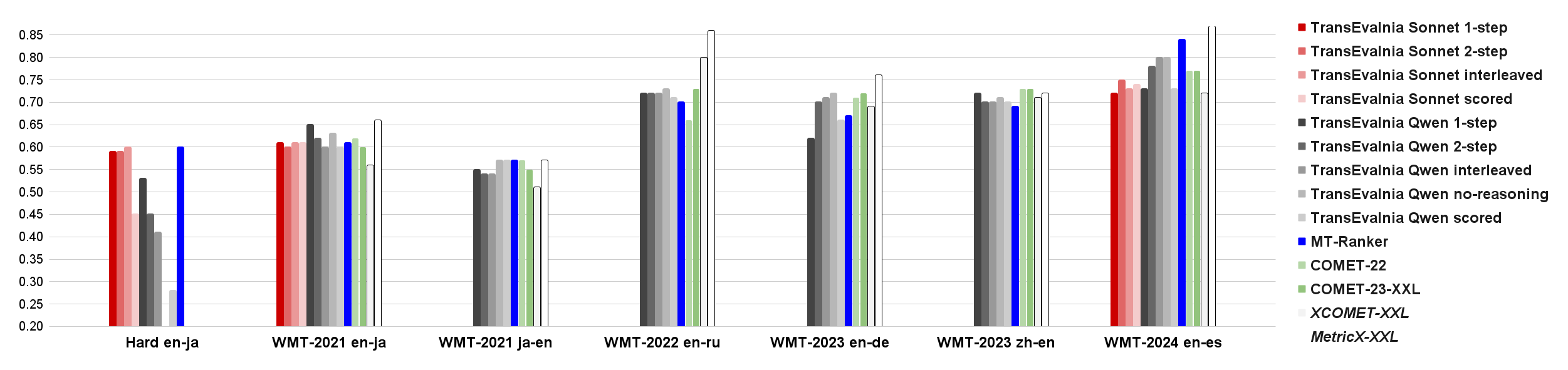}
\end{center}
\caption{\label{fig:accuracies}Accuracies for all systems run on the
  datasets. Sonnet \TransEvalnia\ models are in red and Qwen
  \TransEvalnia\ models in gray.  MT-Ranker is the blue column in each set, the
  two COMET models in shades of green, and XCOMET-XXL and MetricX-XXL are the
  boxed light gray and white columns. Note that XCOMET-XXL and MetricX-XXL are
  fine-tuned on WMT data.  Shown are the means of the two scores for the
  different orders in which the translations were presented to the system. See
  Appendix~\ref{subapp:evaluation_of_ranking_methods} for more fine-grained
  analysis. Sonnet models were only run on Hard~en-ja, WMT-2021~en-ja and
  WMT-2021~en-es. Qwen~no-reasoning was not run on Hard~en-ja.}
\end{figure}

\subsection{Position bias}
\label{subsec:position_bias}

We computed the position bias of ranking a given system as best over our various
corpora. For the Generic and Haiku corpora, we compared the results for three
permutations of the target translations, whereas for the corpora with human
scores, we had two permutations, since in those cases there were only two
translations for any given input. Bias inconsistency is computed as
$B = \sum_{i=1}^{n} \frac{|b_i|}{n}$,
where $n$ is the number of source sentences, and $|b_i|$ is the cardinality of
the set of 'best' translations, which can range from 1 to $p$, the number of
permutations. Obviously then the best (lowest) value of $B$ is 1 and the worst
is $p$.  Within a given system, the interleaved (3-step) variant does indeed
yield the lowest bias inconsistency in the majority (10/14) of cases. For Qwen,
however, in 4 cases the 2-step method yielded the lowest value for $B$.  For the
7 corpora with human ratings, MT-ranker had the lowest value for $B$ in 4 cases,
but 2 were tied with the value for Qwen's interleaved system.

Table~\ref{tab:position_bias_summary} summarizes the above.  See
Tables~\ref{tab:generic}--\ref{tab:wmt-2024_en-es} in
Appendix~\ref{subapp:evaluation_of_position_bias} for detailed data on all of
the experiments.  As documented in those tables, the Qwen~no-reasoning system
used with the WMT data in general had the highest $B$ value among the Qwen-based
variants, with the exception of WMT-2021~ja-en, where the Qwen~1-step system had
a higher value.

\begin{table}[t]
\begin{center}
\begin{tabular}{llrr}
\toprule
Corpus & Best system & Incons. ($\downarrow$) & /N \\
\midrule
Generic & \textbf{\TransEvalnia\ Qwen interleaved} & \textbf{1.47} & /3 \\
Haiku 100 & \textbf{\TransEvalnia\ Sonnet interleaved} & \textbf{1.29} & /3 \\
Haiku Full & \textbf{\TransEvalnia\ Qwen 2-step} & \textbf{1.42} & /3 \\
\midrule
Hard en-ja & \textbf{\TransEvalnia\ Sonnet interleaved} & \textbf{1.04} & /2 \\
WMT-2021 en-ja & \textbf{\TransEvalnia\ Sonnet interleaved} & \textbf{1.15} & /2 \\
WMT-2021 ja-en & \textbf{MT-Ranker} & \textbf{1.14} & /2 \\
WMT-2022 en-ru & \textbf{MT-Ranker}/\textbf{\TransEvalnia\ Qwen interleaved} & \textbf{1.14} & /2 \\
WMT-2023 zh-en & \textbf{MT-Ranker}/\textbf{\TransEvalnia\ Qwen interleaved} & \textbf{1.16} & /2 \\
WMT-2023 en-de & \textbf{\TransEvalnia\ Qwen interleaved} & \textbf{1.11} & /2 \\
WMT-2024 en-es & \textbf{MT-Ranker} & \textbf{1.13} & /2 \\
\bottomrule
\end{tabular}
\end{center}
\caption{\label{tab:position_bias_summary}Summary of lowest position bias
  \emph{inconsistency} results for the various corpora. Note the number of
  permutations tested (\emph{/N}) for each. See
  Appendix~\ref{subapp:evaluation_of_position_bias} for a detailed breakdown of
  results. Note that for the Generic and Haiku datasets MT-Ranker was not
  included and for the Haiku Full only the Qwen-based \TransEvalnia\ systems
  were included.}
\end{table}

\subsection{Human meta-evaluation}
\label{subsec:human_metaeval}

As described in the instructions in Appendix~\ref{subapp:rater_instructions},
the raters were instructed to mark when they \emph{disagree} with the system's
evaluation for a particular part, and to indicate what their disagreement was in
that case.  For both vendors, with Sonnet as the evaluator, agreement with the
fine-grained evaluation is around 0.85, for the overall evaluation around 0.60
for \vendordf\ and 0.69 for \vendorfl.  Agreement on the overall evaluation is
generally higher for Sonnet's evaluations on the small language model translator
(Mistral), but generally lower for the fine-grained
evaluations. In the case where Qwen was used as the evaluator, we had
\vendorfl\ perform the meta-evaluation. The results were broadly similar to Sonnet,
with the overall fine-grained evaluation at 0.85, and the overall evaluation 5
points lower at 0.64. In this case, Mistral's meta-evaluations were worse across
the board. Details can be found in
Appendix~\ref{subapp:human_eval_of_evals}, where we also present results for
meta-evaluation of haiku translations.

\begin{table}[t]
\begin{center}
\begin{tabular}{lrrr}
\toprule
System & \vendordf ($\uparrow$) & \vendorfl ($\uparrow$) & \vendorfl\ (Qwen) ($\uparrow$) \\
\midrule
Sonnet & \textbf{3.84} & \textbf{4.37} & \textbf{4.61}\\
GPT-4o-24 & 3.68 & 4.32 & 4.36\\
Reference & 3.35 & 3.94 & 4.05\\
Qwen & 3.17 & 3.81 & 3.76\\
Mistral & 2.30 & 2.70 & 2.62\\
\bottomrule
\end{tabular}
\end{center}
\caption{\label{tab:overall_scores}Mean overall Likert scores assigned to the
  different translation systems by the two vendors, with Sonnet as the
  evaluator, and for Qwen with \vendorfl, arranged from highest to
  lowest scores.}
\end{table}

Turning now to scoring, the mean assigned Likert score for the overall
translations from each of the systems for the two vendors for Sonnet's
evaluations, and \vendorfl\ for Qwen's evaluations, is shown in
Table~\ref{tab:overall_scores}. Two points are noteworthy. First, the human
reference translation is ranked lower than the two top-performing LLMs, Sonnet
and GPT-4o. This is consistent with results reported elsewhere that as of 2024,
LLM translation systems can outperform human translators
\citep{Freitag:EtAl:24}. Second, both vendors ranked Sonnet as the best
translator in terms of overall quality, though the difference with GPT-4o is
less for \vendorfl~than for \vendordf.  It might seem that a possible confound
is that the evaluations of all translations were also written by Sonnet; Sonnet
may tend to favor its own translations, even though it is not told the source of
the translations. However, the raters from \vendorfl\ also rated the
translations from Sonnet best when Qwen was the evaluator (rightmost
column).

Since our goal is to have raters provide a meta-evaluation of the LLMs'
evaluations of the translations, it was necessary to give them both the
translation and the system outputs, but of course this introduces the danger
that the raters were being influenced by the system evaluations in their
decisions. However, given that the ranking of the translations is largely consistent across
vendors and evaluation systems, this suggests that the raters are not being
unduly influenced by the evaluations in their assessment of the translations
themselves.

Table~\ref{tab:scoring_correlation} shows the Spearman correlations for overall
translation rating score for the two vendors, and for each of the vendors with
Sonnet's rating scores. As can be seen, the correlation for Sonnet is on a par with the
correlations between the two sets of human raters.
\begin{table}[t]
\begin{center}
\begin{tabular}{llrrr}
\toprule
System 1 & System 2 & Spearman's $R$ & Spearman's $R^2$ & $p$\\
\midrule
\vendorfl & \vendordf & 0.52 & 0.27 & 0.00\\
\vendorfl & Sonnet & 0.51 & 0.26 & 0.00\\
\vendordf & Sonnet & 0.54 & 0.29 & 0.00\\
\bottomrule
\end{tabular}
\end{center}
\caption{\label{tab:scoring_correlation}Correlation of Sonnet
  scoring of translations with with those of human evaluators.}
\end{table}
We also ran \TransEvalnia\ to produce evaluations and scores for the dataset using
Qwen and Llama. Details can be found in
Appendix~\ref{subapp:details_of_score_correlations}.


\section{Discussion and future work}
\label{sec:discussion}

In this paper we have presented \TransEvalnia, a prompting-based translation
evaluation and ranking system. The system presents fine-grained evaluations
based on a subset of the MQM, returns an assessment of which translation it
deems the best, and can provide numerical scores for the various dimensions and
for the overall translation. \TransEvalnia\ performs better than the
state-of-the-art MT-Ranker system \citep{Moosa:EtAl:24} on our own data and
several language pairs from various WMT shared tasks. Our system is outperformed
by MetricX-XXL in most cases on the WMT data, but this is likely because
MetricX-XXL has been fine-tuned on WMT. On WMT-2022~en-ru, WMT-2023~en-de and
WMT-2023~zh-en, the system was also outperformed by one or more of the COMET
series rankers. The evaluations returned by Sonnet are deemed highly acceptable
to human raters from our vendors, and the scores for Sonnet and various systems
correlate well with scores assigned by the raters.

Further improvements to open-source models such as Qwen can involve fine-tuning
of those models. For example, we fine-tuned our Qwen model for two epochs using
LoRA \citep{Hu:EtAl:21}, to predict scores on 15,050 Chinese-English and 4,799
German-English translations from WMT-2023 \citep{Blain:EtAl:23}, with associated
MQM ratings. To prepare the data for fine-tuning, the translation pairs were
evaluated and scored using \TransEvalnia, with Qwen as the LLM. The WMT MQM
ratings were converted to a Likert scale for each of the available
dimensions. During fine-tuning, the model was instructed to try to emulate the
scores derived from the WMT data. This resulted in a 5-point improvement on the
correlation for overall scores for English-Japanese with
\vendorfl\ (0.43$\rightarrow$0.48) and a 14-point improvement
(0.34$\rightarrow$0.48) for \vendordf. This suggests that fine-tuning a model
for a translation scoring task can benefit scoring for translations in different
language pairs.

One of the outstanding problems remains position bias with respect to the order
in which the translations are presented, a topic we discuss in detail in
Section~\ref{subsec:reason_eval}. As we have seen, we can often mitigate against
this by decomposing the evaluation into stages, and interleaving the evaluations
of multiple translations, but we cannot entirely eliminate it. As we have also
seen, position bias is also a problem for state-of-the-art ranking systems like
MT-Ranker. As \cite{Wang:EtAl:25:Eliminating} suggest, reasons for position bias in
transformer-based models include positional encodings, which will encode the
same translation and subsequent evaluations differently depending on where they
occur in the input, as well as the causal masking. This partially explains why
interleaving the translation parts and their evaluations may help, since that
will spread the components related to the translations around. Thus no
translation and its evaluations will be wholly within a particular positional
encoding span. We believe further research is needed to try to address the bias
problem from these angles.

\section*{Reproducibility Statement}

All code and data used in this project is open-sourced, enabling others to
reproduce our results.  Depending on the LLM used, results may of course differ.

\section*{Ethics Statement}

Work in Generative AI presents profound ethical concerns, most of which are
well-known, having been discussed at length both in the technical literature and
in the popular press. The work presented here does not seem to present any
additional ethical issues beyond those that are already well-known.



\section*{Acknowledgments}

We thank Tom Gally and Naomi Kagaya for their work on the 'hard'
Japanese-English dataset and for providing expert feedback.

\bibliography{refs}

\begin{thebibliography}{35}
\providecommand{\natexlab}[1]{#1}
\providecommand{\url}[1]{\texttt{#1}}
\expandafter\ifx\csname urlstyle\endcsname\relax
  \providecommand{\doi}[1]{doi: #1}\else
  \providecommand{\doi}{doi: \begingroup \urlstyle{rm}\Url}\fi

\bibitem[Akhbardeh et~al.(2021)Akhbardeh, Arkhangorodsky, Biesialska, Bojar, Chatterjee, Chaudhary, Costa-jussa, Espa{\~n}a-Bonet, Fan, Federmann, Freitag, Graham, Grundkiewicz, Haddow, Harter, Heafield, Homan, Huck, Amponsah-Kaakyire, Kasai, Khashabi, Knight, Kocmi, Koehn, Lourie, Monz, Morishita, Nagata, Nagesh, Nakazawa, Negri, Pal, Tapo, Turchi, Vydrin, and Zampieri]{Akhbardeh:EtAl:21}
Farhad Akhbardeh, Arkady Arkhangorodsky, Magdalena Biesialska, Ond{\v{r}}ej Bojar, Rajen Chatterjee, Vishrav Chaudhary, Marta~R. Costa-jussa, Cristina Espa{\~n}a-Bonet, Angela Fan, Christian Federmann, Markus Freitag, Yvette Graham, Roman Grundkiewicz, Barry Haddow, Leonie Harter, Kenneth Heafield, Christopher Homan, Matthias Huck, Kwabena Amponsah-Kaakyire, Jungo Kasai, Daniel Khashabi, Kevin Knight, Tom Kocmi, Philipp Koehn, Nicholas Lourie, Christof Monz, Makoto Morishita, Masaaki Nagata, Ajay Nagesh, Toshiaki Nakazawa, Matteo Negri, Santanu Pal, Allahsera~Auguste Tapo, Marco Turchi, Valentin Vydrin, and Marcos Zampieri.
\newblock Findings of the 2021 conference on machine translation ({WMT}21).
\newblock In Loic Barrault, Ondrej Bojar, Fethi Bougares, Rajen Chatterjee, Marta~R. Costa-jussa, Christian Federmann, Mark Fishel, Alexander Fraser, Markus Freitag, Yvette Graham, Roman Grundkiewicz, Paco Guzman, Barry Haddow, Matthias Huck, Antonio~Jimeno Yepes, Philipp Koehn, Tom Kocmi, Andre Martins, Makoto Morishita, and Christof Monz (eds.), \emph{Proceedings of the Sixth Conference on Machine Translation}, pp.\  1--88, Online, November 2021. Association for Computational Linguistics.
\newblock URL \url{https://aclanthology.org/2021.wmt-1.1/}.

\bibitem[Ankner et~al.(2024)Ankner, Paul, Cui, Chang, and Ammanabrolu]{Anker:EtAl:24}
Zachary Ankner, Mansheej Paul, Brandon Cui, Jonathan~D. Chang, and Prithviraj Ammanabrolu.
\newblock Critique-out-loud reward models, 2024.
\newblock URL \url{https://arxiv.org/abs/2408.11791}.

\bibitem[Blain et~al.(2023)Blain, Zerva, Rei, Guerreiro, Kanojia, C.~de Souza, Silva, Vaz, Jingxuan, Azadi, Orasan, and Martins]{Blain:EtAl:23}
Frederic Blain, Chrysoula Zerva, Ricardo Rei, Nuno~M. Guerreiro, Diptesh Kanojia, Jos{\'e}~G. C.~de Souza, Beatriz Silva, T{\^a}nia Vaz, Yan Jingxuan, Fatemeh Azadi, Constantin Orasan, and Andr{\'e} Martins.
\newblock Findings of the {WMT} 2023 shared task on quality estimation.
\newblock In Philipp Koehn, Barry Haddow, Tom Kocmi, and Christof Monz (eds.), \emph{Proceedings of the Eighth Conference on Machine Translation}, pp.\  629--653, Singapore, December 2023. Association for Computational Linguistics.
\newblock \doi{10.18653/v1/2023.wmt-1.52}.
\newblock URL \url{https://aclanthology.org/2023.wmt-1.52/}.

\bibitem[Brown et~al.(2020)Brown, Mann, Ryder, Subbiah, Kaplan, Dhariwal, Neelakantan, Shyam, Sastry, Askell, Agarwal, Herbert-Voss, Krueger, Henighan, Child, Ramesh, Ziegler, Wu, Winter, Hesse, Chen, Sigler, Litwin, Gray, Chess, Clark, Berner, McCandlish, Radford, Sutskever, and Amodei]{gpt3}
Tom~B. Brown, Benjamin Mann, Nick Ryder, Melanie Subbiah, Jared Kaplan, Prafulla Dhariwal, Arvind Neelakantan, Pranav Shyam, Girish Sastry, Amanda Askell, Sandhini Agarwal, Ariel Herbert-Voss, Gretchen Krueger, Tom Henighan, Rewon Child, Aditya Ramesh, Daniel~M. Ziegler, Jeffrey Wu, Clemens Winter, Christopher Hesse, Mark Chen, Eric Sigler, Mateusz Litwin, Scott Gray, Benjamin Chess, Jack Clark, Christopher Berner, Sam McCandlish, Alec Radford, Ilya Sutskever, and Dario Amodei.
\newblock Language models are few-shot learners, 2020.
\newblock URL \url{https://arxiv.org/abs/2005.14165}.

\bibitem[Callison-Burch et~al.(2006)Callison-Burch, Osborne, and Koehn]{Callison:EtAl:06}
Chris Callison-Burch, Miles Osborne, and Philipp Koehn.
\newblock Re-evaluating the role of {B}leu in machine translation research.
\newblock In Diana McCarthy and Shuly Wintner (eds.), \emph{11th Conference of the {E}uropean Chapter of the Association for Computational Linguistics}, pp.\  249--256, Trento, Italy, April 2006. Association for Computational Linguistics.
\newblock URL \url{https://aclanthology.org/E06-1032/}.

\bibitem[Conneau et~al.(2020)Conneau, Khandelwal, Goyal, Chaudhary, Wenzek, Guzm{\'a}n, Grave, Ott, Zettlemoyer, and Stoyanov]{XLM}
Alexis Conneau, Kartikay Khandelwal, Naman Goyal, Vishrav Chaudhary, Guillaume Wenzek, Francisco Guzm{\'a}n, Edouard Grave, Myle Ott, Luke Zettlemoyer, and Veselin Stoyanov.
\newblock Unsupervised cross-lingual representation learning at scale.
\newblock In Dan Jurafsky, Joyce Chai, Natalie Schluter, and Joel Tetreault (eds.), \emph{Proceedings of the 58th Annual Meeting of the Association for Computational Linguistics}, pp.\  8440--8451, Online, July 2020. Association for Computational Linguistics.
\newblock \doi{10.18653/v1/2020.acl-main.747}.
\newblock URL \url{https://aclanthology.org/2020.acl-main.747/}.

\bibitem[Fernandes et~al.(2023)Fernandes, Deutsch, Finkelstein, Riley, Martins, Neubig, Garg, Clark, Freitag, and Firat]{Fernandes:EtAl:23}
Patrick Fernandes, Daniel Deutsch, Mara Finkelstein, Parker Riley, Andr{\'e} Martins, Graham Neubig, Ankush Garg, Jonathan Clark, Markus Freitag, and Orhan Firat.
\newblock The devil is in the errors: Leveraging large language models for fine-grained machine translation evaluation.
\newblock In Philipp Koehn, Barry Haddow, Tom Kocmi, and Christof Monz (eds.), \emph{Proceedings of the Eighth Conference on Machine Translation}, pp.\  1066--1083, Singapore, December 2023. Association for Computational Linguistics.
\newblock \doi{10.18653/v1/2023.wmt-1.100}.
\newblock URL \url{https://aclanthology.org/2023.wmt-1.100/}.

\bibitem[Freitag et~al.(2024)Freitag, Mathur, Deutsch, Lo, Avramidis, Rei, Thompson, Blain, Kocmi, Wang, Adelani, Buchicchio, Zerva, and Lavie]{Freitag:EtAl:24}
Markus Freitag, Nitika Mathur, Daniel Deutsch, Chi-Kiu Lo, Eleftherios Avramidis, Ricardo Rei, Brian Thompson, Frederic Blain, Tom Kocmi, Jiayi Wang, David~Ifeoluwa Adelani, Marianna Buchicchio, Chrysoula Zerva, and Alon Lavie.
\newblock Are {LLM}s breaking {MT} metrics? results of the {WMT}24 metrics shared task.
\newblock In Barry Haddow, Tom Kocmi, Philipp Koehn, and Christof Monz (eds.), \emph{Proceedings of the Ninth Conference on Machine Translation}, pp.\  47--81, Miami, Florida, USA, November 2024. Association for Computational Linguistics.
\newblock \doi{10.18653/v1/2024.wmt-1.2}.
\newblock URL \url{https://aclanthology.org/2024.wmt-1.2/}.

\bibitem[Graham et~al.(2013)Graham, Baldwin, Moffat, and Zobel]{Graham:EtAl:13}
Yvette Graham, Timothy Baldwin, Alistair Moffat, and Justin Zobel.
\newblock Continuous measurement scales in human evaluation of machine translation.
\newblock In Antonio Pareja-Lora, Maria Liakata, and Stefanie Dipper (eds.), \emph{Proceedings of the 7th Linguistic Annotation Workshop and Interoperability with Discourse}, pp.\  33--41, Sofia, Bulgaria, August 2013. Association for Computational Linguistics.
\newblock URL \url{https://aclanthology.org/W13-2305/}.

\bibitem[Guerreiro et~al.(2023)Guerreiro, Rei, van Stigt, Coheur, Colombo, and Martins]{Guerreiro:EtAl:23:comet}
Nuno~M. Guerreiro, Ricardo Rei, Daan van Stigt, Luisa Coheur, Pierre Colombo, and André F.~T. Martins.
\newblock xcomet: Transparent machine translation evaluation through fine-grained error detection, 2023.
\newblock URL \url{https://arxiv.org/abs/2310.10482}.

\bibitem[Hu et~al.(2021)Hu, Shen, Wallis, Allen-Zhu, Li, Wang, Wang, and Chen]{Hu:EtAl:21}
Edward~J. Hu, Yelong Shen, Phillip Wallis, Zeyuan Allen-Zhu, Yuanzhi Li, Shean Wang, Lu~Wang, and Weizhu Chen.
\newblock {LoRA}: Low-rank adaptation of large language models, 2021.
\newblock URL \url{https://arxiv.org/abs/2106.09685}.

\bibitem[Juraska et~al.(2023)Juraska, Finkelstein, Deutsch, Siddhant, Mirzazadeh, and Freitag]{Juraska:EtAl:23}
Juraj Juraska, Mara Finkelstein, Daniel Deutsch, Aditya Siddhant, Mehdi Mirzazadeh, and Markus Freitag.
\newblock {M}etric{X}-23: The {G}oogle submission to the {WMT} 2023 metrics shared task.
\newblock In Philipp Koehn, Barry Haddow, Tom Kocmi, and Christof Monz (eds.), \emph{Proceedings of the Eighth Conference on Machine Translation}, pp.\  756--767, Singapore, December 2023. Association for Computational Linguistics.
\newblock \doi{10.18653/v1/2023.wmt-1.63}.
\newblock URL \url{https://aclanthology.org/2023.wmt-1.63/}.

\bibitem[Juraska et~al.(2024)Juraska, Deutsch, Finkelstein, and Freitag]{MetricX24}
Juraj Juraska, Daniel Deutsch, Mara Finkelstein, and Markus Freitag.
\newblock {M}etric{X}-24: The {G}oogle submission to the {WMT} 2024 metrics shared task.
\newblock In Barry Haddow, Tom Kocmi, Philipp Koehn, and Christof Monz (eds.), \emph{Proceedings of the Ninth Conference on Machine Translation}, pp.\  492--504, Miami, Florida, USA, November 2024. Association for Computational Linguistics.
\newblock URL \url{https://aclanthology.org/2024.wmt-1.35}.

\bibitem[Kocmi \& Federmann(2023)Kocmi and Federmann]{GEMBA}
Tom Kocmi and Christian Federmann.
\newblock Large language models are state-of-the-art evaluators of translation quality.
\newblock In Mary Nurminen, Judith Brenner, Maarit Koponen, Sirkku Latomaa, Mikhail Mikhailov, Frederike Schierl, Tharindu Ranasinghe, Eva Vanmassenhove, Sergi~Alvarez Vidal, Nora Aranberri, Mara Nunziatini, Carla~Parra Escart{\'i}n, Mikel Forcada, Maja Popovic, Carolina Scarton, and Helena Moniz (eds.), \emph{Proceedings of the 24th Annual Conference of the European Association for Machine Translation}, pp.\  193--203, Tampere, Finland, June 2023. European Association for Machine Translation.
\newblock URL \url{https://aclanthology.org/2023.eamt-1.19/}.

\bibitem[Kocmi et~al.(2022)Kocmi, Bawden, Bojar, Dvorkovich, Federmann, Fishel, Gowda, Graham, Grundkiewicz, Haddow, Knowles, Koehn, Monz, Morishita, Nagata, Nakazawa, Nov{\'a}k, Popel, and Popovi{\'c}]{Kocmi:EtAl:22}
Tom Kocmi, Rachel Bawden, Ond{\v{r}}ej Bojar, Anton Dvorkovich, Christian Federmann, Mark Fishel, Thamme Gowda, Yvette Graham, Roman Grundkiewicz, Barry Haddow, Rebecca Knowles, Philipp Koehn, Christof Monz, Makoto Morishita, Masaaki Nagata, Toshiaki Nakazawa, Michal Nov{\'a}k, Martin Popel, and Maja Popovi{\'c}.
\newblock Findings of the 2022 conference on machine translation ({WMT}22).
\newblock In Philipp Koehn, Lo{\"i}c Barrault, Ond{\v{r}}ej Bojar, Fethi Bougares, Rajen Chatterjee, Marta~R. Costa-juss{\`a}, Christian Federmann, Mark Fishel, Alexander Fraser, Markus Freitag, Yvette Graham, Roman Grundkiewicz, Paco Guzman, Barry Haddow, Matthias Huck, Antonio Jimeno~Yepes, Tom Kocmi, Andr{\'e} Martins, Makoto Morishita, Christof Monz, Masaaki Nagata, Toshiaki Nakazawa, Matteo Negri, Aur{\'e}lie N{\'e}v{\'e}ol, Mariana Neves, Martin Popel, Marco Turchi, and Marcos Zampieri (eds.), \emph{Proceedings of the Seventh Conference on Machine Translation (WMT)}, pp.\  1--45, Abu Dhabi, United Arab Emirates (Hybrid), December 2022. Association for Computational Linguistics.
\newblock URL \url{https://aclanthology.org/2022.wmt-1.1/}.

\bibitem[Kocmi et~al.(2023)Kocmi, Avramidis, Bawden, Bojar, Dvorkovich, Federmann, Fishel, Freitag, Gowda, Grundkiewicz, Haddow, Koehn, Marie, Monz, Morishita, Murray, Nagata, Nakazawa, Popel, Popovi{\'c}, and Shmatova]{Kocmi:EtAl:23}
Tom Kocmi, Eleftherios Avramidis, Rachel Bawden, Ond{\v{r}}ej Bojar, Anton Dvorkovich, Christian Federmann, Mark Fishel, Markus Freitag, Thamme Gowda, Roman Grundkiewicz, Barry Haddow, Philipp Koehn, Benjamin Marie, Christof Monz, Makoto Morishita, Kenton Murray, Makoto Nagata, Toshiaki Nakazawa, Martin Popel, Maja Popovi{\'c}, and Mariya Shmatova.
\newblock Findings of the 2023 conference on machine translation ({WMT}23): {LLM}s are here but not quite there yet.
\newblock In Philipp Koehn, Barry Haddow, Tom Kocmi, and Christof Monz (eds.), \emph{Proceedings of the Eighth Conference on Machine Translation}, pp.\  1--42, Singapore, December 2023. Association for Computational Linguistics.
\newblock \doi{10.18653/v1/2023.wmt-1.1}.
\newblock URL \url{https://aclanthology.org/2023.wmt-1.1/}.

\bibitem[Kocmi et~al.(2024{\natexlab{a}})Kocmi, Avramidis, Bawden, Bojar, Dvorkovich, Federmann, Fishel, Freitag, Gowda, Grundkiewicz, Haddow, Karpinska, Koehn, Marie, Monz, Murray, Nagata, Popel, Popovi{\'c}, Shmatova, Steingr{\'i}msson, and Zouhar]{Kocmi:EtAl:2024}
Tom Kocmi, Eleftherios Avramidis, Rachel Bawden, Ond{\v{r}}ej Bojar, Anton Dvorkovich, Christian Federmann, Mark Fishel, Markus Freitag, Thamme Gowda, Roman Grundkiewicz, Barry Haddow, Marzena Karpinska, Philipp Koehn, Benjamin Marie, Christof Monz, Kenton Murray, Masaaki Nagata, Martin Popel, Maja Popovi{\'c}, Mariya Shmatova, Steinth{\'o}r Steingr{\'i}msson, and Vil{\'e}m Zouhar.
\newblock Findings of the {WMT}24 general machine translation shared task: The {LLM} era is here but {MT} is not solved yet.
\newblock In Barry Haddow, Tom Kocmi, Philipp Koehn, and Christof Monz (eds.), \emph{Proceedings of the Ninth Conference on Machine Translation}, pp.\  1--46, Miami, Florida, USA, November 2024{\natexlab{a}}. Association for Computational Linguistics.
\newblock \doi{10.18653/v1/2024.wmt-1.1}.
\newblock URL \url{https://aclanthology.org/2024.wmt-1.1/}.

\bibitem[Kocmi et~al.(2024{\natexlab{b}})Kocmi, Zouhar, Federmann, and Post]{Kocmi:EtAl:24}
Tom Kocmi, Vilém Zouhar, Christian Federmann, and Matt Post.
\newblock Navigating the metrics maze: Reconciling score magnitudes and accuracies, 2024{\natexlab{b}}.
\newblock URL \url{https://arxiv.org/abs/2401.06760}.

\bibitem[Li et~al.(2024)Li, Wang, Ma, Wu, Wang, Gao, and Liu]{Li:EtAl:24}
Zongjie Li, Chaozheng Wang, Pingchuan Ma, Daoyuan Wu, Shuai Wang, Cuiyun Gao, and Yang Liu.
\newblock Split and merge: Aligning position biases in {LLM}-based evaluators, 2024.
\newblock URL \url{https://arxiv.org/abs/2310.01432}.

\bibitem[Liu et~al.(2025)Liu, Lyu, Wu, Wang, Luo, Zhang, and Shang]{Liu:EtAl:2025}
Sinuo Liu, Chenyang Lyu, Minghao Wu, Longyue Wang, Weihua Luo, Kaifu Zhang, and Zifu Shang.
\newblock New trends for modern machine translation with large reasoning models, 2025.
\newblock URL \url{https://arxiv.org/abs/2503.10351}.

\bibitem[Mathur et~al.(2020)Mathur, Baldwin, and Cohn]{Mathur-EtAl-20}
Nitika Mathur, Timothy Baldwin, and Trevor Cohn.
\newblock Tangled up in {BLEU}: Reevaluating the evaluation of automatic machine translation evaluation metrics.
\newblock In Dan Jurafsky, Joyce Chai, Natalie Schluter, and Joel Tetreault (eds.), \emph{Proceedings of the 58th Annual Meeting of the Association for Computational Linguistics}, pp.\  4984--4997, Online, July 2020. Association for Computational Linguistics.
\newblock \doi{10.18653/v1/2020.acl-main.448}.
\newblock URL \url{https://aclanthology.org/2020.acl-main.448/}.

\bibitem[Moosa et~al.(2024)Moosa, Zhang, and Yin]{Moosa:EtAl:24}
Ibraheem~Muhammad Moosa, Rui Zhang, and Wenpeng Yin.
\newblock Mt-ranker: Reference-free machine translation evaluation by inter-system ranking, 2024.
\newblock URL \url{https://arxiv.org/abs/2401.17099}.

\bibitem[Papineni et~al.(2002)Papineni, Roukos, Ward, and Zhu]{Papineni:EtAl:02}
Kishore Papineni, Salim Roukos, Todd Ward, and Wei-Jing Zhu.
\newblock {B}leu: a method for automatic evaluation of machine translation.
\newblock In Pierre Isabelle, Eugene Charniak, and Dekang Lin (eds.), \emph{Proceedings of the 40th Annual Meeting of the Association for Computational Linguistics}, pp.\  311--318, Philadelphia, Pennsylvania, USA, July 2002. Association for Computational Linguistics.
\newblock \doi{10.3115/1073083.1073135}.
\newblock URL \url{https://aclanthology.org/P02-1040/}.

\bibitem[Rei et~al.(2022)Rei, Treviso, Guerreiro, Zerva, Farinha, Maroti, C.~de Souza, Glushkova, Alves, Coheur, Lavie, and Martins]{CometKiwi}
Ricardo Rei, Marcos Treviso, Nuno~M. Guerreiro, Chrysoula Zerva, Ana~C Farinha, Christine Maroti, Jos{\'e}~G. C.~de Souza, Taisiya Glushkova, Duarte Alves, Luisa Coheur, Alon Lavie, and Andr{\'e} F.~T. Martins.
\newblock {C}omet{K}iwi: {IST}-unbabel 2022 submission for the quality estimation shared task.
\newblock In Philipp Koehn, Lo{\"i}c Barrault, Ond{\v{r}}ej Bojar, Fethi Bougares, Rajen Chatterjee, Marta~R. Costa-juss{\`a}, Christian Federmann, Mark Fishel, Alexander Fraser, Markus Freitag, Yvette Graham, Roman Grundkiewicz, Paco Guzman, Barry Haddow, Matthias Huck, Antonio Jimeno~Yepes, Tom Kocmi, Andr{\'e} Martins, Makoto Morishita, Christof Monz, Masaaki Nagata, Toshiaki Nakazawa, Matteo Negri, Aur{\'e}lie N{\'e}v{\'e}ol, Mariana Neves, Martin Popel, Marco Turchi, and Marcos Zampieri (eds.), \emph{Proceedings of the Seventh Conference on Machine Translation (WMT)}, pp.\  634--645, Abu Dhabi, United Arab Emirates (Hybrid), December 2022. Association for Computational Linguistics.
\newblock URL \url{https://aclanthology.org/2022.wmt-1.60/}.

\bibitem[Sellam et~al.(2020)Sellam, Das, and Parikh]{BLEURT}
Thibault Sellam, Dipanjan Das, and Ankur Parikh.
\newblock {BLEURT}: Learning robust metrics for text generation.
\newblock In Dan Jurafsky, Joyce Chai, Natalie Schluter, and Joel Tetreault (eds.), \emph{Proceedings of the 58th Annual Meeting of the Association for Computational Linguistics}, pp.\  7881--7892, Online, July 2020. Association for Computational Linguistics.
\newblock \doi{10.18653/v1/2020.acl-main.704}.
\newblock URL \url{https://aclanthology.org/2020.acl-main.704/}.

\bibitem[Shi et~al.(2024)Shi, Ma, Liang, Ma, and Vosoughi]{Shi:EtAl:24}
Lin Shi, Chiyu Ma, Wenhua Liang, Weicheng Ma, and Soroush Vosoughi.
\newblock Judging the judges: A systematic study of position bias in {LLM}-as-a-judge, 2024.
\newblock URL \url{https://arxiv.org/abs/2406.07791}.

\bibitem[Song et~al.(2025)Song, Riley, Deutsch, and Freitag]{Song:EtAl:25}
Yixiao Song, Parker Riley, Daniel Deutsch, and Markus Freitag.
\newblock Enhancing human evaluation in machine translation with comparative judgment, 2025.
\newblock URL \url{https://arxiv.org/abs/2502.17797}.

\bibitem[Wang et~al.(2025{\natexlab{a}})Wang, Pham, Moghimifar, and Vu]{Wang:EtAl:25}
Minghan Wang, Viet-Thanh Pham, Farhad Moghimifar, and Thuy-Trang Vu.
\newblock Proverbs run in pairs: Evaluating proverb translation capability of large language model, 2025{\natexlab{a}}.
\newblock URL \url{https://arxiv.org/abs/2501.11953}.

\bibitem[Wang et~al.(2025{\natexlab{b}})Wang, Zhang, Li, Huang, Han, Ji, Kakade, Peng, and Ji]{Wang:EtAl:25:Eliminating}
Ziqi Wang, Hanlin Zhang, Xiner Li, Kuan-Hao Huang, Chi Han, Shuiwang Ji, Sham~M. Kakade, Hao Peng, and Heng Ji.
\newblock Eliminating position bias of language models: A mechanistic approach.
\newblock In \emph{The Thirteenth International Conference on Learning Representations}, 2025{\natexlab{b}}.
\newblock URL \url{https://openreview.net/forum?id=fvkElsJOsN}.

\bibitem[Xu et~al.(2023)Xu, Wang, Pan, Song, Freitag, Wang, and Li]{Xu:EtAl:23}
Wenda Xu, Danqing Wang, Liangming Pan, Zhenqiao Song, Markus Freitag, William Wang, and Lei Li.
\newblock {INSTRUCTSCORE}: Towards explainable text generation evaluation with automatic feedback.
\newblock In Houda Bouamor, Juan Pino, and Kalika Bali (eds.), \emph{Proceedings of the 2023 Conference on Empirical Methods in Natural Language Processing}, pp.\  5967--5994, Singapore, December 2023. Association for Computational Linguistics.
\newblock \doi{10.18653/v1/2023.emnlp-main.365}.
\newblock URL \url{https://aclanthology.org/2023.emnlp-main.365/}.

\bibitem[Xu et~al.(2024)Xu, Deutsch, Finkelstein, Juraska, Zhang, Liu, Wang, Li, and Freitag]{LLMRefine}
Wenda Xu, Daniel Deutsch, Mara Finkelstein, Juraj Juraska, Biao Zhang, Zhongtao Liu, William~Yang Wang, Lei Li, and Markus Freitag.
\newblock {LLMR}efine: Pinpointing and refining large language models via fine-grained actionable feedback.
\newblock In Kevin Duh, Helena Gomez, and Steven Bethard (eds.), \emph{Findings of the Association for Computational Linguistics: NAACL 2024}, pp.\  1429--1445, Mexico City, Mexico, June 2024. Association for Computational Linguistics.
\newblock \doi{10.18653/v1/2024.findings-naacl.92}.
\newblock URL \url{https://aclanthology.org/2024.findings-naacl.92/}.

\bibitem[Xue et~al.(2021)Xue, Constant, Roberts, Kale, Al-Rfou, Siddhant, Barua, and Raffel]{mT5}
Linting Xue, Noah Constant, Adam Roberts, Mihir Kale, Rami Al-Rfou, Aditya Siddhant, Aditya Barua, and Colin Raffel.
\newblock m{T}5: A massively multilingual pre-trained text-to-text transformer.
\newblock In Kristina Toutanova, Anna Rumshisky, Luke Zettlemoyer, Dilek Hakkani-Tur, Iz~Beltagy, Steven Bethard, Ryan Cotterell, Tanmoy Chakraborty, and Yichao Zhou (eds.), \emph{Proceedings of the 2021 Conference of the North American Chapter of the Association for Computational Linguistics: Human Language Technologies}, pp.\  483--498, Online, June 2021. Association for Computational Linguistics.
\newblock \doi{10.18653/v1/2021.naacl-main.41}.
\newblock URL \url{https://aclanthology.org/2021.naacl-main.41/}.

\bibitem[Yan et~al.(2024)Yan, Liu, Chiu, Shen, Qin, Yu, Lakshmanan, Kurzion, Rush, Liu, and Bendersky]{Yan:EtAl:24}
Jing~Nathan Yan, Tianqi Liu, Justin Chiu, Jiaming Shen, Zhen Qin, Yue Yu, Charumathi Lakshmanan, Yair Kurzion, Alexander Rush, Jialu Liu, and Michael Bendersky.
\newblock Predicting text preference via structured comparative reasoning.
\newblock In Lun-Wei Ku, Andre Martins, and Vivek Srikumar (eds.), \emph{Proceedings of the 62nd Annual Meeting of the Association for Computational Linguistics (Volume 1: Long Papers)}, pp.\  10040--10060, Bangkok, Thailand, August 2024. Association for Computational Linguistics.
\newblock \doi{10.18653/v1/2024.acl-long.541}.
\newblock URL \url{https://aclanthology.org/2024.acl-long.541/}.

\bibitem[Ye et~al.(2024)Ye, Wang, Huang, Chen, Zhang, Moniz, Gao, Geyer, Huang, Chen, Chawla, and Zhang]{Ye:EtAl:24}
Jiayi Ye, Yanbo Wang, Yue Huang, Dongping Chen, Qihui Zhang, Nuno Moniz, Tian Gao, Werner Geyer, Chao Huang, Pin-Yu Chen, Nitesh~V Chawla, and Xiangliang Zhang.
\newblock Justice or prejudice? quantifying biases in {LLM}-as-a-judge, 2024.
\newblock URL \url{https://arxiv.org/abs/2410.02736}.

\bibitem[Zeng et~al.(2025)Zeng, He, Ren, Liang, Xiao, Zhou, Sun, and Yu]{Zeng:EtEl:25}
Jie Zeng, Qianyu He, Qingyu Ren, Jiaqing Liang, Yanghua Xiao, Weikang Zhou, Zeye Sun, and Fei Yu.
\newblock Order matters: Investigate the position bias in multi-constraint instruction following, 2025.
\newblock URL \url{https://arxiv.org/abs/2502.17204}.

\end{thebibliography}
\bibliographystyle{asstyle}

\newpage
\appendix
\section{Appendix}

\subsection{Details of data}
\label{subapp:data}

\subsubsection{Data without human rankings}
\label{subsubsec:data_noranking}

\paragraph{Eigokotozawa and NTREX.}

Eigokotozawa is a collection of 862 English proverbs with reference
translations into Japanese, for which we also generated automated translations
using the following systems:
\begin{itemize}
\item
  Mistral
\item
  Qwen
\item
  Sonnet
\item
  GPT-4o-24
\end{itemize}
Source: \url{https://eigokotozawa.net}.

NTREX has a collection of 1,997 English sentences from news sources with
Japanese translations. As above, we generated four LLM translations with the
aforenamed systems. Source: \url{https://github.com/MicrosoftTranslator/NTREX}.

We selected 50 examples of an English source and its Japanese translations, from
each of the above sets and combined them into a dataset which we will henceforth
term \textbf{Generic}. The motivation is to provide a set of translations that
ranges in difficulty from news style texts---relatively easy to translate; to
proverbs---harder to translate, see e.g. \citep{Wang:EtAl:25}.  We used samples
from this Generic set for human meta-evaluation of our automatic evaluations
conducted by third party vendors.

\paragraph{Haiku.}

1,065 haiku by Matsuo Bash\=o (1644--1694).
Source:
\url{https://www2.yamanashi-ken.ac.jp/~itoyo/basho/haikusyu/Default.htm}.
We generated LLM translations using Mistral, Qwen, Sonnet and GPT-4 (rather than
GPT-4o-24 as used above).  From these we generated a random subset of 100 with
the original Japanese source and all translations
(henceforth \textbf{Haiku~100}), in addition to using the full set
(\textbf{Haiku~Full}).

\subsubsection{Data with human rankings}
\label{subsubsec:data_ranking}

\paragraph{English-to-Japanese `hard' set}

Two expert Japanese-English translators produced a set of 10 English source
sentences from which we produced multiple translations using various MT systems
and LLMs (Google Translate, GPT 3.5, GPT 4, and GPT with a reasoning phase). The
original English sentences were designed to be tricky to render into
Japanese. The experts then rated the translations on a 10-point scale, with 10.0
being the best. We will refer to this dataset as \textbf{Hard~en-ja}.

\paragraph{WMT-2021.}

For WMT~2021 \citep{Akhbardeh:EtAl:21} (Source:
\url{https://www.statmt.org/wmt21/}), we used news texts from the
\textbf{Japanese-English}/\textbf{English-Japanese}
(\textbf{WMT-2021~ja-en}/\textbf{WMT-2021~en-ja}) portion consisting of 1,851
sentences translated by various systems. For each direction, we constructed 500
randomly selected triples consisting of a source-language text and two
target-language translations, along with their ratings. Human scores used Direct
Assessment \citep{Graham:EtAl:13}.  Of the two translations we arbitrarily
designated the first as ``reference'' and the other as ``other'': Note that
these are just arbitrary terms, since in general neither of the translations is
an actual reference translation in this dataset.

\paragraph{WMT-2022.}

We used the \textbf{English-Russian} portions of WMT~2022 \citep{Kocmi:EtAl:22}
(Source: \url{https://www.statmt.org/wmt22/}) (\textbf{WMT-2022~en-ru}),
consisting of a mix of 2,016 news, E-commerce and other texts. As with WMT~2021
data, we constructed 500 randomly selected triples consisting of a
source-language text and two target-language translations, along with their
ratings. All translations had human-rated MQM scores. For ranking the two
translations we used the overall MQM score, where a lower value is better, as
the human rating for ranking purposes.

\paragraph{WMT-2023.}

We used the \textbf{Chinese-English} (\textbf{WMT-2023~zh-en}) and
\textbf{English-German} (\textbf{WMT-2023~en-de}) portions of the WMT~2023
shared task \citep{Kocmi:EtAl:23,Blain:EtAl:23} (Source:
\url{https://www2.statmt.org/wmt23/}). The Chinese-English portion consisted of
18,832 sentence translation pairs, which were a mix of news, user manuals and
E-commerce texts. The English-German portion had 5,980 pairs, consisting of
social media, news, meeting notes and E-commerce texts. All translations had
human-rated MQM scores. From these sets, for each language pair, we constructed
a random subset of 500 translation triples consisting of one source-language
sentence, and two target-language translations. As before, of the two we
arbitrarily designated the first as ``reference'' and the other as ``other''.

\paragraph{WMT-2024.}

We used the \textbf{English-Spanish} (\textbf{WMT-2024~en-es}) portions of
WMT~2024 \citep{Kocmi:EtAl:24} (Source: \url{https://www2.statmt.org/wmt24/}),
which had hand-labeled MQM scores. Construction of our subset proceeded along
the same lines as with previous WMT sets.

\subsection{Single-step Rating Prompt}
\label{subapp:single_step}

All prompts given in this and the following sections are specific to the
English-Japanese case. In the actual version used, language-pair-specific
versions of the one-shot examples are given, conditioned on the
\verb|{{source-language}}|-\verb|{{target-language}}| variables.

\begin{promptbox}{Single-step rating prompt}{Prompts/single_step_rating.txt}
\end{promptbox}

\subsection{Two-Step Prompts}
\label{subapp:two_step}

\subsubsection{Evaluation Prompt}
\label{subsubapp:evaluation_prompt}

\begin{promptbox}{Evaluation prompt}{Prompts/evaluation.txt}
\end{promptbox}

\subsubsection{Ranking Prompt}
\label{subsubapp:ranking_prompt}

\begin{promptbox}{Ranking prompt}{Prompts/ranking.txt}
\end{promptbox}

\subsection{Three-step (interleaving) prompts}
\label{subapp:three_step}

The evaluation prompt is as in Appendix~\ref{subsubapp:evaluation_prompt} above.
The instructions for interleaving and ranking are as follows:

\subsubsection{Interleaving prompt}
\label{subsubapp:interleaving}

\begin{promptbox}{Interleaving prompt}{Prompts/interleaving.txt}
\end{promptbox}

\subsubsection{Interleaved ranking prompt}
\label{subsubapp:interleaved_ranking}

\begin{promptbox}{Interleaved ranking prompt}{Prompts/interleaved_ranking.txt}
\end{promptbox}

\subsection{Scoring prompt}
\label{subapp:scoring_prompt}

\begin{promptbox}{Scoring prompt}{Prompts/scoring.txt}
\end{promptbox}

\subsection{No-reasoning ranking prompt}
\label{subapp:no_reasoning}

\begin{promptbox}{No-reasoning ranking prompt}{Prompts/no_reasoning.txt}
\end{promptbox}

\subsection{Instructions for Raters}
\label{subapp:rater_instructions}

\begin{promptbox}{Instructions for raters}{Prompts/rater_instructions.txt}
\end{promptbox}

\subsection{Example of three-step interleaving method.}
\label{subapp:basho_three_step}

\begin{figure}[t]
\begin{center}
\includegraphics[width=1.0\textwidth]{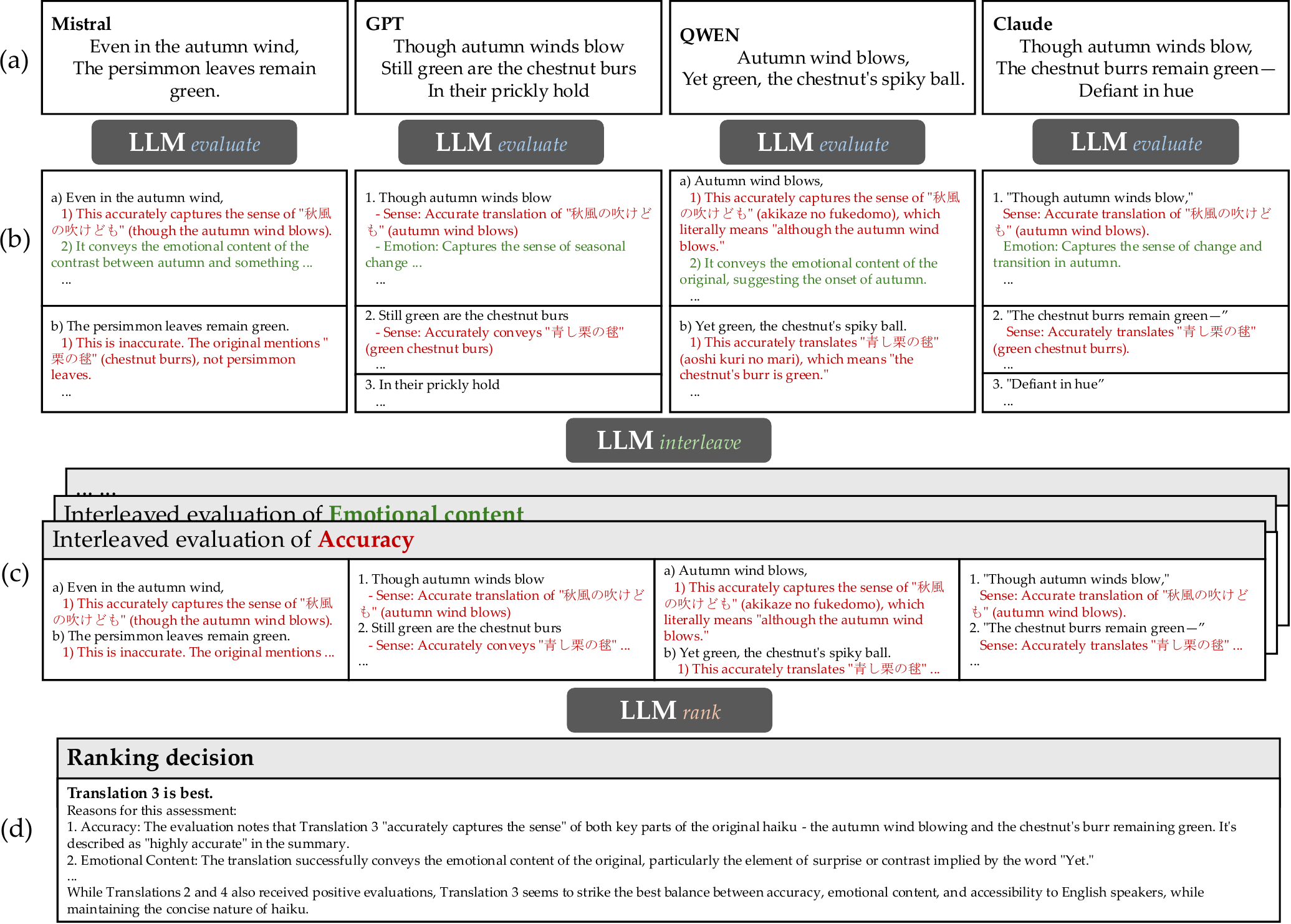}
\end{center}
\caption{\label{fig:haiku_eval}Sample evaluation of a Matsuo Bash\={o} haiku
  秋風の吹けども青し栗の毬
  \textit{akikaze no, fuke domo aoshi, kuri no iga}, `Though
  the autumn wind blows, the chestnut burrs are yet green', with four
  machine-generated translations (a). The ranking system (Sonnet)
  first evaluated each translation (b), then interleaved these evaluations
  (c). Finally on the basis of the interleaved translations, it ranked the
  translations, giving its reasons for the ranking, picking the Qwen translation
  as the best (d).}
\end{figure}

Figure~\ref{fig:haiku_eval} shows an example of translations of a Matsuo
Bash\=o haiku and their evaluation and ranking using the three-step method.
Note that the prompts for the haiku case are slightly different from the ones
used for the generic text, reflecting the somewhat different evaluation
dimensions discussed above.\footnote{Full versions of all prompts used will be
released along with the software.}

\subsection{Evaluation of ranking methods}
\label{subapp:evaluation_of_ranking_methods}

Results for individual corpora for ranking accuracy can be found in
Tables~\ref{tab:hard_en_ja}--\ref{tab:wmt_en_es}.

In the tables, the first line ``First
Best'' indicates the accuracy one would get with the dataset if one always
picked the first translation as the best.  For each system, we present the
accuracies when the first translation is in fact presented first---indicated as
\texttt{SystemName/1}---and when the order of the two translations is
flipped---\texttt{SystemName/2}. As can be seen, these two permutations can
result in a quite large spread, indicating position bias in the ranking
system. For example in the Hard en-ja set (Table~\ref{tab:hard_en_ja}), we can
see that MT-Ranker has a strong bias towards the first translation, which in
this particular set is always the right answer, so that MT-Ranker/1 has an
accuracy of 0.74, whereas MT-Ranker/2 has an accuracy of 0.47. On the other
hand, we can see that for \TransEvalnia, both Qwen~1-step and Sonnet~1-step have
a strong second-position bias.

Comparisons were also run against COMET-22, COMET-23-XXL, XCOMET-XXL and
MetricX-XXL. In most cases we mark in bold-face the system with the best
performance. Due to the variance of the results within a presentation order
permutation, we determine the best system to be the one with the
highest \emph{minimum} value across the two permutations. In the case of a tie
for the highest minimum, we break the tie by also considering the highest
maximum. This is a more fine-grained assessment than the mean score displayed in
Figure~\ref{fig:accuracies}, but is arguably more meaningful since merely taking
the mean of the two scores obscures sometimes significant differences due to
position bias. However for XCOMET-XXL and MetricX-XXL, since these have been
fine-tuned on WMT data, we cannot be sure that results for these systems are not
biased. In cases where those systems got the highest score, this is indicated in
italics.

Note that with the exception of WMT-2021~ja-en, where it achieves the best
accuracy, the scoring-based ranking is inferior to other methods. Thus, while
this method has the advantage that the scoring is done separately for each
translation, and thus is not affected by the position bias problem, it is less
desirable in terms of its overall performance.

\begin{table}[t]
\begin{center}
\begin{tabular}{rrrr}
\toprule
\multicolumn{4}{c}{ Hard en-ja } \\
\midrule
System & \# Cor & N & Acc $\uparrow$ \\
\midrule
First best & 47 & 47 & 1.00 \\
MT-Ranker/1 & 34 & 47 & 0.72 \\
MT-Ranker/2 & 22 & 47 & 0.47 \\
Qwen 1-step/1 & 23 & 47 & 0.49 \\
Qwen 1-step/2 & 27 & 47 & 0.57 \\
Qwen 2-step/1 & 22 & 47 & 0.47 \\
Qwen 2-step/2 & 20 & 47 & 0.43 \\
Qwen interleaved/1 & 17 & 47 & 0.36 \\
Qwen interleaved/2 & 22 & 47 & 0.47 \\
Sonnet 1-step/1 & 20 & 47 & 0.43 \\
Sonnet 1-step/2 & 35 & 47 & 0.74 \\
Sonnet 2-step/1 & 27 & 47 & 0.57 \\
Sonnet 2-step/2 & 28 & 47 & 0.60 \\
\textbf{Sonnet interleaved/1} & 28 & 47 & \textbf{0.60} \\
\textbf{Sonnet interleaved/2} & 28 & 47 & \textbf{0.60} \\
Qwen scored & 13 & 47 & 0.28 \\
Sonnet scored & 21 & 47 & 0.45 \\
\bottomrule
\end{tabular}
\end{center}
\caption{\label{tab:hard_en_ja}System accuracies: Hard en-ja}
\end{table}

\begin{table}[t]
\begin{center}
\begin{tabular}{rrrr}
\toprule
\multicolumn{4}{c}{ WMT-2021 en-ja } \\
\midrule
System & \# Cor & N & Acc $\uparrow$ \\
\midrule
First best & 256 & 500 & 0.51 \\
MT-Ranker/1 & 311 & 500 & 0.62 \\
MT-Ranker/2 & 300 & 500 & 0.60 \\
COMET-22 & 309 & 500 & 0.62 \\
COMET-23-XXL & 300 & 500 & 0.60 \\
XCOMET-XXL & 281 & 500 & 0.56 \\
\textit{MetricX-XXL} & 331 & 500 & \textit{0.66} \\
Qwen no-reasoning/1 & 325 & 500 & 0.65 \\
Qwen no-reasoning/2 & 307 & 500 & 0.61 \\
\textbf{Qwen 1-step/1} & 325 & 500 & \textbf{0.65} \\
\textbf{Qwen 1-step/2} & 321 & 500 & \textbf{0.64} \\
Qwen 2-step/1 & 315 & 500 & 0.63 \\
Qwen 2-step/2 & 308 & 500 & 0.62 \\
Qwen interleaved/1 & 302 & 500 & 0.60 \\
Qwen interleaved/2 & 302 & 500 & 0.60 \\
Sonnet 1-step/1 & 307 & 500 & 0.61 \\
Sonnet 1-step/2 & 301 & 500 & 0.60 \\
Sonnet 2-step/1 & 305 & 500 & 0.61 \\
Sonnet 2-step/2 & 294 & 500 & 0.59 \\
Sonnet interleaved/1 & 308 & 500 & 0.62 \\
Sonnet interleaved/2 & 307 & 500 & 0.61 \\
Qwen scored & 302 & 500 & 0.60 \\
Sonnet scored & 304 & 500 & 0.61 \\
\bottomrule
\end{tabular}
\end{center}
\caption{\label{tab:wmt_en_ja}System accuracies: WMT-2021 en-ja}
\end{table}

\begin{table}[t]
\begin{center}
\begin{tabular}{rrrr}
\toprule
\multicolumn{4}{c}{ WMT-2021 ja-en } \\
\midrule
System & \# Cor & N & Acc $\uparrow$ \\
\midrule
First best & 246 & 500 & 0.49 \\
MT-Ranker/1 & 289 & 500 & 0.58 \\
MT-Ranker/2 & 285 & 500 & 0.57 \\
COMET-22 & 287 & 500 & 0.57 \\
COMET-23-XXL & 277 & 500 & 0.55 \\
XCOMET-XXL & 256 & 500 & 0.51 \\
\textit{MetricX-XXL} & 284 & 500 & \textit{0.57} \\
Qwen no-reasoning/1 & 292 & 500 & 0.58 \\
Qwen no-reasoning/2 & 275 & 500 & 0.55 \\
Qwen 1-step/1 & 276 & 500 & 0.55 \\
Qwen 1-step/2 & 271 & 500 & 0.54 \\
Qwen 2-step/1 & 273 & 500 & 0.55 \\
Qwen 2-step/2 & 271 & 500 & 0.54 \\
Qwen interleaved/1 & 271 & 500 & 0.54 \\
Qwen interleaved/2 & 267 & 500 & 0.53 \\
\textbf{Qwen scored} & 286 & 500 & \textbf{0.57} \\
\bottomrule
\end{tabular}
\end{center}
\caption{\label{tab:wmt_ja_en}System accuracies: WMT-2021 ja-en}
\end{table}

\begin{table}[t]
\begin{center}
\begin{tabular}{rrrr}
\toprule
\multicolumn{4}{c}{ WMT-2022 en-ru } \\
\midrule
System & \# Cor & N & Acc $\uparrow$ \\
\midrule
First best & 298 & 500 & 0.60 \\
MT-Ranker/1 & 350 & 500 & 0.70 \\
MT-Ranker/2 & 354 & 500 & 0.71 \\
COMET-22 & 332 & 500 & 0.66 \\
\textbf{COMET-23-XXL} & 363 & 500 & \textbf{0.73} \\
XCOMET-XXL & 401 & 500 & 0.80 \\
\textit{MetricX-XXL} & 428 & 500 & \textit{0.86} \\
Qwen no-reasoning/1 & 352 & 500 & 0.70 \\
Qwen no-reasoning/2 & 375 & 500 & 0.75 \\
Qwen 1-step/1 & 350 & 500 & 0.70 \\
Qwen 1-step/2 & 367 & 500 & 0.73 \\
Qwen 2-step/1 & 355 & 500 & 0.71 \\
Qwen 2-step/2 & 368 & 500 & 0.74 \\
Qwen interleaved/1 & 354 & 500 & 0.71 \\
Qwen interleaved/2 & 366 & 500 & 0.73 \\
Qwen scored & 353 & 500 & 0.71 \\
\bottomrule
\end{tabular}
\end{center}
\caption{\label{tab:wmt_en_ru}System accuracies: WMT-2022 en-ru}
\end{table}

\begin{table}[t]
\begin{center}
\begin{tabular}{rrrr}
\toprule
\multicolumn{4}{c}{ WMT-2023 en-de } \\
\midrule
System & \# Cor & N & Acc $\uparrow$ \\
\midrule
First best & 196 & 500 & 0.39 \\
MT-Ranker/1 & 329 & 500 & 0.66 \\
MT-Ranker/2 & 344 & 500 & 0.69 \\
COMET-22 & 354 & 500 & 0.71 \\
\textbf{COMET-23-XXL} & 362 & 500 & \textbf{0.72} \\
XCOMET-XXL & 344 & 500 & 0.69 \\
\textit{MetricX-XXL} & 378 & 500 & \textit{0.76} \\
Qwen no-reasoning/1 & 363 & 500 & 0.73 \\
Qwen no-reasoning/2 & 355 & 500 & 0.71 \\
Qwen 1-step/1 & 316 & 500 & 0.63 \\
Qwen 1-step/2 & 301 & 500 & 0.60 \\
Qwen 2-step/1 & 343 & 500 & 0.69 \\
Qwen 2-step/2 & 358 & 500 & 0.72 \\
Qwen interleaved/1 & 356 & 500 & 0.71 \\
Qwen interleaved/2 & 357 & 500 & 0.71 \\
Qwen scored & 330 & 500 & 0.66 \\
\bottomrule
\end{tabular}
\end{center}
\caption{\label{tab:wmt_en_de}System accuracies: WMT-2023 en-de}
\end{table}

\begin{table}[t]
\begin{center}
\begin{tabular}{rrrr}
\toprule
\multicolumn{4}{c}{ WMT-2023 zh-en } \\
\midrule
System & \# Cor & N & Acc $\uparrow$ \\
\midrule
First best & 262 & 500 & 0.52 \\
MT-Ranker/1 & 348 & 500 & 0.70 \\
MT-Ranker/2 & 347 & 500 & 0.69 \\
\textbf{COMET-22} & 365 & 500 & \textbf{0.73} \\
\textbf{COMET-23-XXL} & 367 & 500 & \textbf{0.73} \\
XCOMET-XXL & 356 & 500 & 0.71 \\
MetricX-XXL & 359 & 500 & 0.72 \\
Qwen no-reasoning/1 & 364 & 500 & 0.73 \\
Qwen no-reasoning/2 & 344 & 500 & 0.69 \\
Qwen 1-step/1 & 355 & 500 & 0.71 \\
Qwen 1-step/2 & 367 & 500 & 0.73 \\
Qwen 2-step/1 & 348 & 500 & 0.70 \\
Qwen 2-step/2 & 353 & 500 & 0.71 \\
Qwen interleaved/1 & 356 & 500 & 0.71 \\
Qwen interleaved/2 & 347 & 500 & 0.69 \\
Qwen scored & 350 & 500 & 0.70 \\
\bottomrule
\end{tabular}
\end{center}
\caption{\label{tab:wmt_zh_en}System accuracies: WMT-2023 zh-en}
\end{table}

\begin{table}[t]
\begin{center}
\begin{tabular}{rrrr}
\toprule
\multicolumn{4}{c}{ WMT-2024 en-es } \\
\midrule
System & \# Cor & N & Acc $\uparrow$ \\
\midrule
First best & 261 & 500 & 0.52 \\
\textbf{MT-Ranker/1} & 420 & 500 & \textbf{0.84} \\
\textbf{MT-Ranker/2} & 425 & 500 & \textbf{0.85} \\
COMET-22 & 387 & 500 & 0.77 \\
COMET-23-XXL & 386 & 500 & 0.77 \\
XCOMET-XXL & 362 & 500 & 0.72 \\
\textit{MetricX-XXL} & 434 & 500 & \textit{0.87} \\
Qwen no-reasoning/1 & 411 & 500 & 0.82 \\
Qwen no-reasoning/2 & 393 & 500 & 0.79 \\
Qwen 1-step/1 & 365 & 500 & 0.73 \\
Qwen 1-step/2 & 367 & 500 & 0.73 \\
Qwen 2-step/1 & 384 & 500 & 0.77 \\
Qwen 2-step/2 & 395 & 500 & 0.79 \\
Qwen interleaved/1 & 403 & 500 & 0.81 \\
Qwen interleaved/2 & 395 & 500 & 0.79 \\
Sonnet 1-step/1 & 360 & 500 & 0.72 \\
Sonnet 1-step/2 & 360 & 500 & 0.72 \\
Sonnet 2-step/1 & 374 & 500 & 0.75 \\
Sonnet 2-step/2 & 372 & 500 & 0.74 \\
Sonnet interleaved/1 & 361 & 500 & 0.72 \\
Sonnet interleaved/2 & 367 & 500 & 0.73 \\
Qwen scored & 363 & 500 & 0.73 \\
Sonnet scored & 372 & 500 & 0.74 \\
\bottomrule
\end{tabular}
\end{center}
\caption{\label{tab:wmt_en_es}System accuracies: WMT-2024 en-es}
\end{table}

\subsection{Evaluation of position bias}
\label{subapp:evaluation_of_position_bias}

Results for individual corpora for position bias can be found in
Tables~\ref{tab:generic}--\ref{tab:wmt-2024_en-es}.  In the tables, the lowest
overall value of $B$ and the associated system is marked in boldface. Within a
given system---say the 3 runs of Qwen---the lowest value is marked in italics if
that is not also the lowest \emph{overall} value.

Within a given system, the interleaved (3-step) variant does indeed yield the
lowest bias inconsistency in the majority (10/14) cases. For Qwen, however, in 4
cases the 2-step method yielded the lowest value for $B$.  For the 7 corpora
with human ratings, MT-ranker had the lowest value for $B$ in 4 cases, but two
of those were tied with the value for Qwen's interleaved system.

\begin{table}[t]
\begin{center}
\begin{tabular}{rrr}
\toprule
\multicolumn{3}{c}{ Generic } \\
\midrule
System & Inconsistency ($\downarrow$) & /N permutations \\
\midrule
Sonnet 1-step & 1.87 & /3 \\
Sonnet 2-step & 1.56 & /3 \\
\textit{Sonnet interleaved} & \textit{1.54} & /3 \\
\midrule
Qwen 1-step & 1.78 & /3 \\
Qwen 2-step & 1.60 & /3 \\
\textbf{Qwen interleaved} & \textbf{1.47} & /3 \\
\bottomrule
\end{tabular}
\end{center}
\caption{\label{tab:generic}Position bias inconsistency for Generic.}
\end{table}

\begin{table}[t]
\begin{center}
\begin{tabular}{rrr}
\toprule
\multicolumn{3}{c}{ Haiku 100 } \\
\midrule
System & Inconsistency ($\downarrow$) & /N permutations \\
\midrule
Sonnet 1-step & 1.85 & /3 \\
Sonnet 2-step & 1.47 & /3 \\
\textbf{Sonnet interleaved} & \textbf{1.29} & /3 \\
\midrule
Qwen 1-step & 1.93 & /3 \\
\textit{Qwen 2-step} & \textit{1.42} & /3 \\
Qwen interleaved & 1.54 & /3 \\
\bottomrule
\end{tabular}
\end{center}
\caption{\label{tab:haiku_100}Position bias inconsistency for Haiku 100.}
\end{table}

\begin{table}[t]
\begin{center}
\begin{tabular}{rrr}
\toprule
\multicolumn{3}{c}{ Haiku Full } \\
\midrule
System & Inconsistency ($\downarrow$) & /N permutations \\
\midrule
Qwen 1-step & 1.80 & /3 \\
\textbf{Qwen 2-step} & \textbf{1.42} & /3 \\
Qwen interleaved & 1.43 & /3 \\
\bottomrule
\end{tabular}
\end{center}
\caption{\label{tab:haiku_full}Position bias inconsistency for Haiku Full.}
\end{table}

\begin{table}[t]
\begin{center}
\begin{tabular}{rrr}
\toprule
\multicolumn{3}{c}{ Hard en-ja } \\
\midrule
System & Inconsistency ($\downarrow$) & /N permutations \\
\midrule
MT-Ranker & 1.26 & /2 \\
\midrule
Sonnet 1-step & 1.32 & /2 \\
Sonnet 2-step & 1.06 & /2 \\
\textbf{Sonnet interleaved} & \textbf{1.04} & /2 \\
\midrule
Qwen 1-step & 1.13 & /2 \\
\textit{Qwen 2-step} & \textit{1.09} & /2 \\
Qwen interleaved & 1.15 & /2 \\
\bottomrule
\end{tabular}
\end{center}
\caption{\label{tab:hard_en-ja}Position bias inconsistency for Hard en-ja.}
\end{table}

\begin{table}[t]
\begin{center}
\begin{tabular}{rrr}
\toprule
\multicolumn{3}{c}{ WMT-2021 en-ja } \\
\midrule
System & Inconsistency ($\downarrow$) & /N permutations \\
\midrule
MT-Ranker & 1.23 & /2 \\
\midrule
Sonnet 1-step & 1.38 & /2 \\
Sonnet 2-step & 1.18 & /2 \\
\textbf{Sonnet interleaved} & \textbf{1.15} & /2 \\
\midrule
Qwen no-reasoning & 1.34 & /2 \\
Qwen 1-step & 1.26 & /2 \\
Qwen 2-step & 1.19 & /2 \\
\textit{Qwen interleaved} & \textit{1.18} & /2 \\
\bottomrule
\end{tabular}
\end{center}
\caption{\label{tab:wmt-2021_en-ja}Position bias inconsistency for WMT-2021 en-ja.}
\end{table}

\begin{table}[t]
\begin{center}
\begin{tabular}{rrr}
\toprule
\multicolumn{3}{c}{ WMT-2021 ja-en } \\
\midrule
System & Inconsistency ($\downarrow$) & /N permutations \\
\midrule
\textbf{MT-Ranker} & \textbf{1.14} & /2 \\
\midrule
Qwen no-reasoning & 1.28 & /2 \\
Qwen 1-step & 1.31 & /2 \\
\textit{Qwen 2-step} & \textit{1.16} & /2 \\
Qwen interleaved & 1.18 & /2 \\
\bottomrule
\end{tabular}
\end{center}
\caption{\label{tab:wmt-2021_ja-en}Position bias inconsistency for WMT-2021 ja-en.}
\end{table}

\begin{table}[t]
\begin{center}
\begin{tabular}{rrr}
\toprule
\multicolumn{3}{c}{ WMT-2022 en-ru } \\
\midrule
System & Inconsistency ($\downarrow$) & /N permutations \\
\midrule
\textbf{MT-Ranker} & \textbf{1.16} & /2 \\
\midrule
Qwen no-reasoning & 1.32 & /2 \\
Qwen 1-step & 1.22 & /2 \\
Qwen 2-step & 1.16 & /2 \\
\textbf{Qwen interleaved} & \textbf{1.16} & /2 \\
\bottomrule
\end{tabular}
\end{center}
\caption{\label{tab:wmt-2022_en-ru}Position bias inconsistency for WMT-2022 en-ru.}
\end{table}

\begin{table}[t]
\begin{center}
\begin{tabular}{rrr}
\toprule
\multicolumn{3}{c}{ WMT-2023 zh-en } \\
\midrule
System & Inconsistency ($\downarrow$) & /N permutations \\
\midrule
\textbf{MT-Ranker} & \textbf{1.11} & /2 \\
\midrule
Qwen no-reasoning & 1.25 & /2 \\
Qwen 1-step & 1.20 & /2 \\
Qwen 2-step & 1.13 & /2 \\
\textbf{Qwen interleaved} & \textbf{1.11} & /2 \\
\bottomrule
\end{tabular}
\end{center}
\caption{\label{tab:wmt-2023_zh-en}Position bias inconsistency for WMT-2023 zh-en.}
\end{table}

\begin{table}[t]
\begin{center}
\begin{tabular}{rrr}
\toprule
\multicolumn{3}{c}{ WMT-2023 en-de } \\
\midrule
System & Inconsistency ($\downarrow$) & /N permutations \\
\midrule
MT-Ranker & 1.27 & /2 \\
\midrule
Qwen no-reasoning & 1.22 & /2 \\
Qwen 1-step & 1.21 & /2 \\
Qwen 2-step & 1.17 & /2 \\
\textbf{Qwen interleaved} & \textbf{1.11} & /2 \\
\bottomrule
\end{tabular}
\end{center}
\caption{\label{tab:wmt-2023_en-de}Position bias inconsistency for WMT-2023 en-de.}
\end{table}

\begin{table}[t]
\begin{center}
\begin{tabular}{rrr}
\toprule
\multicolumn{3}{c}{ WMT-2024 en-es } \\
\midrule
System & Inconsistency ($\downarrow$) & /N permutations \\
\midrule
\textbf{MT-Ranker} & \textbf{1.13} & /2 \\
\midrule
Qwen no-reasoning & 1.20 & /2 \\
Qwen 1-step & 1.17 & /2 \\
Qwen 2-step & 1.16 & /2 \\
\textit{Qwen interleaved} & \textit{1.14} & /2 \\
\bottomrule
\end{tabular}
\end{center}
\caption{\label{tab:wmt-2024_en-es}Position bias inconsistency for WMT-2024 en-es.}
\end{table}

\subsection{Human evaluation of \TransEvalnia's evaluations}
\label{subapp:human_eval_of_evals}

Tables~\ref{tab:meta_eval_vendordf} and \ref{tab:meta_eval_vendorfl} present
the results for the meta-evaluation between the human raters and Sonnet's
evaluations of the translations for the Generic dataset, broken down by
translator, and span---fine-grained and overall. As described in the
instructions in Appendix~\ref{subapp:rater_instructions}, the raters were
instructed to mark when they \emph{disagree} with the system's evaluation for a
particular part, and to indicate what their disagreement is in that case.  For
both vendors, agreement with the fine-grained evaluation is around 0.85, for the
overall evaluation around 0.60 for \vendordf\ and 0.69
for \vendorfl. Agreement on the overall evaluation is generally higher for
Sonnet's evaluations on the small language model translator (Mistral), but
generally lower for the fine-grained evaluations.

Table~\ref{tab:meta_eval_vendorfl_qwen} presents similar results from \vendorfl,
for evaluations using Qwen as the evaluator. While the Qwen model is weaker than
Sonnet, on balance the correlation with human raters is only somewhat lower,
with fine-grained evaluation 0.85, identical with the ratings from \vendorfl\
for Sonnet, and 0.64 versus 0.69 for the overall evaluation, a 5-point drop.

In Table~\ref{tab:meta_eval_vendorfl_qwen_haiku} we present meta-evaluations
from \vendorfl\ with Qwen as the evaluator for the harder task of evaluating
translations of Matsuo Bash\=o haiku into English. In this case the vendor
employed a different rater, one who is natively bilingual in English and
Japanese, and is deeply familiar with both Japanese and American
culture. Agreement with the evaluations is lower than for the English-Japanese
translations of news texts and proverbs, which is expected given that
translating and evaluating translations of poetry is a harder task, but the
agreement with the fine-grained evaluations still reached 0.79. For this task we
modified the evaluation criteria slightly: ACCURACY and TERMINOLOGY were
combined into the criterion of whether the translation preserved the
SENSE~OF~THE~ORIGINAL, and an additional category of EMOTIONAL~CONTENT was
added, which is needed for poetry. The overall meta-evaluation agreement on this
dimension was 0.75, suggesting that the TransEvalnia evaluation on this
difficult dimension was reasonable three quarters of the time.

As an example of a disagreement with Sonnet's evaluation of a translation,
consider the following case where raters from both vendors agreed that the
evaluation was problematic.  The example involves GPT-4o-24's translation of
\begin{quote}
SNL started the show with a skit starring Matt Damon in which the Hollywood star
made fun of Brett Kavanaugh's testimony before the Senate Judicial Committee on
sexual assault claims made by Christine Blasey Ford.
\end{quote}
as
\begin{quote}
\underline{SNLは、マット・デイモンが主演するスキットでショーを開始しました。}このスキットで
は、ハリウッドスターがクリスティン・ブレイジー・フォードによる性的暴行の主張に関
するブレット・カバノーの上院司法委員会での証言をからかいました。
\end{quote}
and in particular the underlined portion above. Sonnet evaluates this on the
ACCURACY dimension as follows:
\begin{quote}
Accurate translation of ``SNL started the show with a skit starring Matt Damon''
\end{quote}
\vendordf's rater comments:
\begin{quote}
This translation is inappropriate for a Japanese audience. SNL is not commonly
recognized as the TV program ``Saturday Night Live.'' Additionally, the
translations for ``show'' and ``skit'' do not cater to the audience effectively.
\end{quote}
In similar fashion, \vendorfl's rater observes:
\begin{quote}
As for the word ``skit'', ``スキット'' is the loanword. ``寸劇'' is more commonly used
and understood in Japanese.
\end{quote}
Complete examples of the cases where the raters disagreed with the LLM's
evaluations can be found in the released data.

\begin{table}[t]
\begin{center}
\begin{tabular}{rrr}
\toprule
Translator & Spans & Agreement \\
\midrule
GPT-4o & Fine-grained & 0.88 \\
 & Overall & 0.51 \\
Reference & Fine-grained & 0.89 \\
 & Overall & 0.68 \\
Sonnet & Fine-grained & 0.91 \\
 & Overall & 0.67 \\
Mistral & Fine-grained & 0.83 \\
 & Overall & 0.68 \\
Qwen & Fine-grained & 0.84 \\
 & Overall & 0.48 \\
\midrule
All translators & Fine-grained & 0.86 \\
 & Overall & 0.60 \\
\bottomrule
\end{tabular}
\end{center}
\caption{\label{tab:meta_eval_vendordf}Human meta-evaluation by \vendordf\ of Sonnet's evaluation for
the Generic (en-ja) corpus, broken down by original translator, and for all
translators combined.}
\end{table}

\begin{table}[t]
\begin{center}
\begin{tabular}{rrr}
\toprule
Translator & Spans & Agreement \\
\midrule
GPT-4o & Fine-grained & 0.91 \\
 & Overall & 0.71 \\
Reference & Fine-grained & 0.84 \\
 & Overall & 0.65 \\
Sonnet & Fine-grained & 0.87 \\
 & Overall & 0.74 \\
Mistral & Fine-grained & 0.78 \\
 & Overall & 0.78 \\
Qwen & Fine-grained & 0.85 \\
 & Overall & 0.55 \\
\midrule
All translators & Fine-grained & 0.85 \\
 & Overall & 0.69 \\
\bottomrule
\end{tabular}
\end{center}
\caption{\label{tab:meta_eval_vendorfl}Human meta-evaluation by \vendorfl\ of Sonnet's evaluation for
the Generic (en-ja) corpus, broken down by original translator, and for all
translators combined.}
\end{table}

\begin{table}[t]
\begin{center}
\begin{tabular}{rrr}
\toprule
Translator & Spans & Agreement \\
\midrule
GPT-4o & Fine-grained & 0.88 \\
 & Overall & 0.69 \\
Reference & Fine-grained & 0.90 \\
 & Overall & 0.73 \\
Sonnet & Fine-grained & 0.86 \\
 & Overall & 0.68 \\
Mistral & Fine-grained & 0.79 \\
 & Overall & 0.53 \\
Qwen & Fine-grained & 0.82 \\
 & Overall & 0.59 \\
\midrule
All translators & Fine-grained & 0.85 \\
 & Overall & 0.64 \\
\bottomrule
\end{tabular}
\end{center}
\caption{\label{tab:meta_eval_vendorfl_qwen}Human meta-evaluation by \vendorfl\ of Qwen's evaluation for
the Generic (en-ja) corpus, broken down by original translator, and for all
translators combined.}
\end{table}

\begin{table}[t]
\begin{center}
\begin{tabular}{rrr}
\toprule
Translator & Spans & Agreement \\
\midrule
GPT-4o & Fine-grained & 0.76 \\
 & Overall & 0.43 \\
Sonnet & Fine-grained & 0.84 \\
 & Overall & 0.52 \\
Mistral & Fine-grained & 0.76 \\
 & Overall & 0.50 \\
Qwen & Fine-grained & 0.81 \\
 & Overall & 0.54 \\
All translators & Fine-grained & 0.79 \\
 & Overall & 0.50 \\
\bottomrule
\end{tabular}
\end{center}
\caption{\label{tab:meta_eval_vendorfl_qwen_haiku}Human meta-evaluation by \vendorfl\ of Qwen's evaluation for
LLM translations from the Haiku (ja-en) corpus, broken down by original translator, and for all
translators combined. In this case a single rater performed all the ratings.}
\end{table}

\subsection{Details on correlations of scores with human scores}
\label{subapp:details_of_score_correlations}

Correlations between both vendors and between all models predicted scores for
Sonnet's evaluations are presented in Table~\ref{tab:combined_spans}. Sonnet
produces the best correlations, on a par with human-human correlation, Qwen
coming in second and Llama a distant third.

\begin{table}[t]
\begin{center}
\begin{tabular}{ll|rrr|rrr}
\toprule
\multirow{2}{*}{System 1} & \multirow{2}{*}{System 2} & \multicolumn{3}{c|}{All spans (/197)} & \multicolumn{3}{c}{Overall span (/65)} \\
\cmidrule(lr){3-5} \cmidrule(lr){6-8}
& & $R$ & $R^2$ & $p$ & $R$ & $R^2$ & $p$ \\
\midrule
\vendorfl & \vendordf & 0.49 & 0.24 & 0.00 & 0.52 & 0.27 & 0.00 \\
\vendorfl & Sonnet & 0.49 & 0.24 & 0.00 & 0.51 & 0.26 & 0.00 \\
\vendorfl & Qwen & 0.37 & 0.14 & 0.00 & 0.43 & 0.19 & 0.00 \\
\vendorfl & Llama & 0.15 & 0.02 & 0.04 & 0.23 & 0.05 & 0.07 \\
\vendordf & Sonnet & 0.46 & 0.21 & 0.00 & 0.54 & 0.29 & 0.00 \\
\vendordf & Qwen & 0.30 & 0.09 & 0.00 & 0.34 & 0.11 & 0.01 \\
\vendordf & Llama & 0.20 & 0.04 & 0.01 & 0.36 & 0.13 & 0.00 \\
Sonnet & Qwen & 0.57 & 0.33 & 0.00 & 0.61 & 0.37 & 0.00 \\
Sonnet & Llama & 0.37 & 0.14 & 0.00 & 0.50 & 0.25 & 0.00 \\
Qwen & Llama & 0.26 & 0.07 & 0.00 & 0.35 & 0.12 & 0.00 \\
\bottomrule
\end{tabular}
\end{center}
\caption{\label{tab:combined_spans}Spearman's correlations between system pairs
across all spans (/197) and overall span (/65), with Sonnet as the evaluator.}
\end{table}

\end{document}